\documentclass[lettersize,journal]{IEEEtran}
\PassOptionsToPackage{table}{xcolor}
\usepackage{amsmath,amsfonts}
\usepackage{array}
\usepackage{orcidlink}
\usepackage[caption=false,font=normalsize,labelfont=sf,textfont=sf]{subfig}
\usepackage{textcomp}
\usepackage{stfloats}
\usepackage{url}
\usepackage{verbatim}
\usepackage{graphicx}
\usepackage{framed}
\usepackage{cite}

\usepackage{pifont}
\newcommand{\XSolidBrush}{\ding{55}}  
\newcommand{\Checkmark}{\ding{51}}   
\definecolor{darkred}{rgb}{0.6,0.0,0.0}
\definecolor{darkblue}{rgb}{0.0,0.0,0.55}
\usepackage{multirow}
\usepackage{booktabs}
\usepackage{colortbl}
\usepackage{float}
\usepackage[ruled,vlined,linesnumbered]{algorithm2e}
\SetKwInput{KwIn}{Input}
\SetKwInput{KwOut}{Output}

\begin{document}

\title{ERIS: An Energy-Guided Feature Disentanglement Framework for Out-of-Distribution Time Series Classification}

\author{
Xin Wu\textsuperscript{\orcidlink{0009-0008-2292-4534}}, 
Fei Teng\IEEEauthorrefmark{1}\textsuperscript{\orcidlink{0000-0001-9535-7245}}, 
Ji Zhang\textsuperscript{\orcidlink{0000-0001-6949-3673}}, 
Xingwang Li\textsuperscript{\orcidlink{0009-0007-9650-130X}}, 
and Yuxuan Liang\textsuperscript{\orcidlink{0000-0003-2817-7337}} \IEEEmembership{Member,~IEEE}
\thanks{
Xin Wu, Fei Teng (the corresponding author), Ji Zhang and Xingwang Li are with the School of Computing and Artificial Intelligence, Southwest Jiaotong University, Chengdu, Sichuan 611756, China. Fei Teng is also with the Engineering Research Center of Sustainable Urban Intelligent Transportation, Ministry of Education, Chengdu, Sichuan 611756, China. (Email: xinwu5386@gmail.com; fteng@swjtu.edu.cn; jizhang.jim@gmail.com; xingwangli98@163.com.)
}
\thanks{Yuxuan Liang is with the Hong Kong University of Science and Technology,
Guangzhou 511458, China (e-mail: yuxliang@outlook.com).}
}

\markboth{Journal of \LaTeX\ Class Files,~Vol.~14, No.~8, August~2021}%
{Shell \MakeLowercase{\textit{et al.}}: A Sample Article Using IEEEtran.cls for IEEE Journals}

\maketitle

\begin{abstract}

An ideal time series classification (TSC) should be able to capture invariant representations, but achieving reliable performance on out-of-distribution (OOD) data remains a core obstacle. This obstacle arises from the way models inherently entangle domain-specific and label-relevant features, resulting in spurious correlations. While feature disentanglement aims to solve this, current methods are largely unguided, lacking the semantic direction required to isolate truly universal features. To address this, we propose an end-to-end \underline{\textbf{E}}nergy-\underline{\textbf{R}}egularized \underline{\textbf{I}}nformation for \underline{\textbf{S}}hift-Robustness (\textbf{ERIS}) framework to enable guided and reliable feature disentanglement. The core idea is that effective disentanglement requires not only mathematical constraints but also semantic guidance to anchor the separation process. ERIS incorporates three key mechanisms to achieve this goal. Specifically, we first introduce an energy-guided calibration mechanism, which provides crucial semantic guidance for the separation, enabling the model to self-calibrate. Additionally, a weight-level orthogonality strategy enforces structural independence between domain-specific and label-relevant features, thereby mitigating their interference. Moreover, an auxiliary adversarial generalization mechanism enhances robustness by injecting structured perturbations. Experiments across four benchmarks demonstrate that ERIS achieves a statistically significant improvement over state-of-the-art baselines, consistently securing the top performance rank.
\end{abstract}


\section{Introduction}

\IEEEPARstart{G}{eneralization} of data-driven models under distribution shifts is a paramount challenge in time series classification (TSC) \cite{Zhou2023KDD, zhang2025Reliable}. For instance, a model proficiently trained to recognize running patterns from a young athlete's sensor data may completely fail when applied to an elderly person, due to significant shifts in the underlying data distribution. This out-of-distribution (OOD) problem poses a direct threat to the reliability of TSC, a cornerstone technology vital across a broad range of scientific and industrial fields \cite{Wei2006KDD,Cao2013tkde,Ffulcher2014rkde,Lines2022KDD}.

For time series data, this problem is particularly severe, as models tend to learn spurious shortcuts between domain-specific and label-relevant features. In human activity recognition, for instance, a model might erroneously associate the high signal amplitude of an athlete with the act of running itself, an entanglement that prevents it from learning a universal activity pattern and instead creates a biased representation \cite{Zhang2023KDD,Chang2024ICDE}. When the data distribution shifts to a different demographic, such as an elderly person, decision boundaries built on these spurious correlations inevitably fail \cite{Liu2023tkde,Xiong2025tkde,Tamang2025tkde}.

\begin{figure}
\centering
\includegraphics[width=0.9\columnwidth]{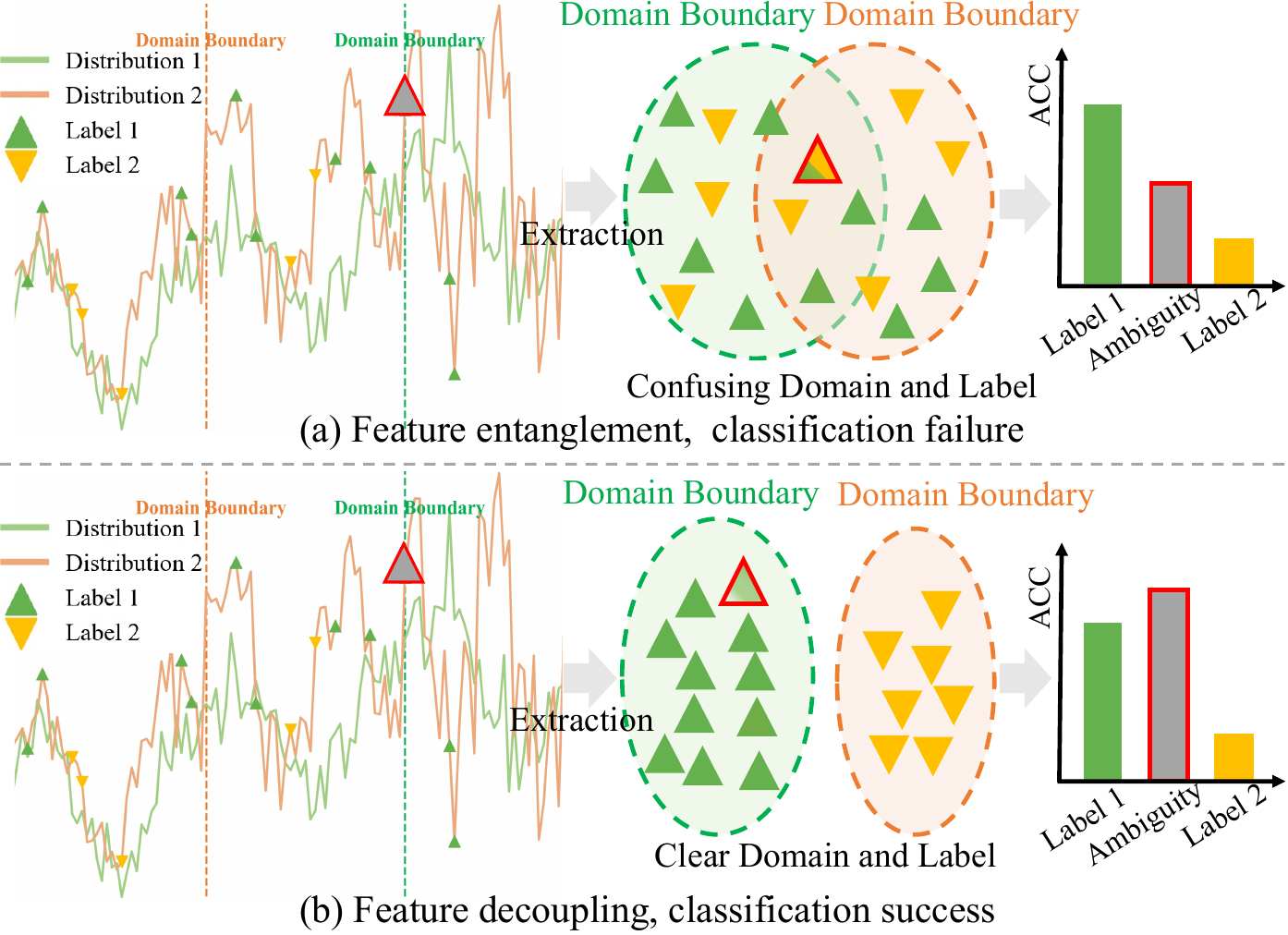} 
\caption{Illustration of domain and label decoupling impact on classification. (a) Coupled features lead to entangled representations, causing poor label separation and reduced accuracy. (b) Decoupled features enable invariant representations, improving label separation and performance.}
\label{fig_intro}
\end{figure}

\begin{figure}
\centering
\includegraphics[width=0.75\columnwidth]{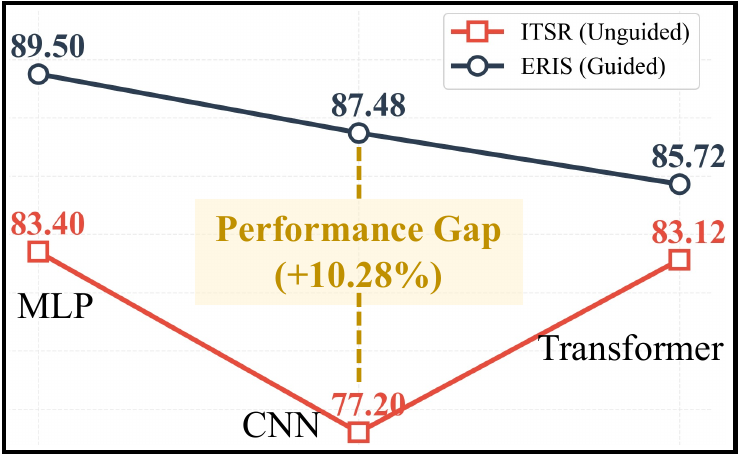}
\caption{A comparative analysis of performance gaps between unguided (ITSR) and guided (ERIS) methods using different model backbones.}
\label{fig_gap}
\end{figure}

Although feature disentanglement aims to solve this, current methods are largely unguided, lacking the semantic direction required to isolate truly universal features \cite{Li22KDD,Liang2024KDD}. These approaches typically impose mathematical constraints like orthogonality, forcing features into separate groups but failing to inform the model which features should be separated \cite{Lee2024VLDB}. This is akin to asking a model to separate  individual-specific patterns from activity-specific patterns without providing any clues to distinguish between them. Consequently, the model can easily converge to a trivial solution that satisfies the mathematical constraints but offers no meaningful OOD generalization. Effective disentanglement, therefore, requires not only mathematical constraints but also semantic guidance to anchor the separation process \cite{Venkateswaran2021KDD}. The term semantic here refers to an interpretable signal correlated with domain shifts, such as the physical energy of the signal. For instance, an athlete's movements possess far greater energy than an elderly person's, making energy a powerful semantic clue. Our core idea is to leverage this energy information to guide the disentanglement process. This approach directs the model to identify and separate features highly correlated with energy variations (i.e., domain-specific attributes), thereby preserving the core, energy-agnostic features of the activity itself. Conceptually, this unguided entanglement leads to the ambiguous boundaries shown in Fig. \ref{fig_intro} (a), in sharp contrast to the ideal, clearly disentangled clusters in Fig. \ref{fig_intro} (b). The unguided ITSR method provides a clear example of this failure: it exhibits a significant performance gap (see Fig. \ref{fig_gap}), and its feature correlation matrix reveals a superficial "checkerboard" pattern (see Fig. \ref{fig_orl}), confirming that its features remain semantically entangled. This observation raises a research question of paramount importance: 
\begin{framed}
{
\textit{How can guided disentanglement be achieved to facilitate a reliable decomposition into domain-specific and label-relevant features?}
}
\end{framed}
Answering this question is fundamental to achieving reliable and generalizable performance in TSC. In this work, we propose an \underline{\textbf{E}}nergy-\underline{\textbf{R}}egularized \underline{\textbf{I}}nformation for \underline{\textbf{S}}hift-Robustness (\textbf{ERIS}) framework to enable guided and reliable feature disentanglement for TSC. The philosophy of energy-based models (EBMs) \cite{lecun2006tutorial} guides the design of ERIS, which circumvents the often intractable task of learning a normalized probability distribution by instead learning a direct energy function to score data compatibility. Critically for time series, the physical energy (such as spectral energy or variance) is domain-specific and closely related to the label \cite{li2024human}. The core idea of ERIS is that effective disentanglement requires not only mathematical constraints but also such semantic guidance that can anchor the separation process.

To this end, ERIS incorporates three key mechanisms. \underline{\textbf{Firstly}}, an energy-guided calibration mechanism leverages this principle to directly address unguided disentanglement by learning to assign low energy to well-aligned features and high energy to poorly aligned ones. More importantly, by linking these energy scores to model uncertainty, ERIS becomes self-calibrating. When confronted with OOD samples that trigger high-energy responses, it automatically lowers confidence, thereby suppressing overconfident, incorrect predictions. \underline{\textbf{Secondly}}, a weight-level orthogonality strategy is employed to enforce structural independence between domain-specific and label-relevant representations, thereby mitigating mutual interference. \underline{\textbf{Thirdly}}, an auxiliary adversarial training mechanism injects structured domain perturbations into the feature space to encourage the learning of invariant, disentangled features. 

The main contributions of this work are threefold:
\begin{itemize}
	\item We demonstrate that excessive entanglement between domain-specific and label-relevant features hinders the model's ability to learn invariant representations.
       \item We propose the ERIS framework, which imposes effective weight-level orthogonality as a mathematical constraint and incorporates energy-based guidance to reliably disentangle domain-specific and label-relevant features.
      \item   A domain-disentangled adversarial training mechanism is designed to mitigate structured perturbations that reflect domain discrepancies during training.
\end{itemize}

\section{Related Work}

\subsection{OOD in Time Series Classification}

OOD generalization is a key problem in reliable AI \cite{Wang2023KDDTrustworthy}. Existing methods fall into two main aspects. The first is robust optimization. It improves resistance to distribution shifts by minimizing worst-case risk. Typical examples include GroupDRO \cite{Sagawa2020Distributionally} and VREx \cite{krueger2021}. The second is invariant representation learning. It aims to extract causal or stable features across domains. Representative methods include IRM \cite{arjovsky2019}. Recent works such as SDL \cite{ye2023}, FOOGD \cite{liao2024}, and EVIL \cite{huang2025} build on these paradigms with new extensions. 

However, directly applying these methods to TSC brings new challenges. Time series data are often non-stationary. They may contain long-term dependencies and concept drift \cite{wang2022generalizing, wu2025}. Current approaches mainly follow two directions. One is distribution alignment. It reduces domain gaps, as seen in AdaRNN \cite{du2021} and Diversify \cite{lu2023}. The other is invariant feature separation. For example, GILE \cite{qian2021} separates core dynamics from temporal styles. ITSR \cite{shi2024} uses information theory to isolate label-relevant features. Many of these methods rely on internal constraints like orthogonality or independence. These constraints may not match the true data semantics. The learned invariance may not generalize beyond the training data. This leads to unreliable results under unseen distributions. A new mechanism is needed to guide models toward meaningful and stable representations.

\begin{figure*}
\centering
\includegraphics[scale=0.475]{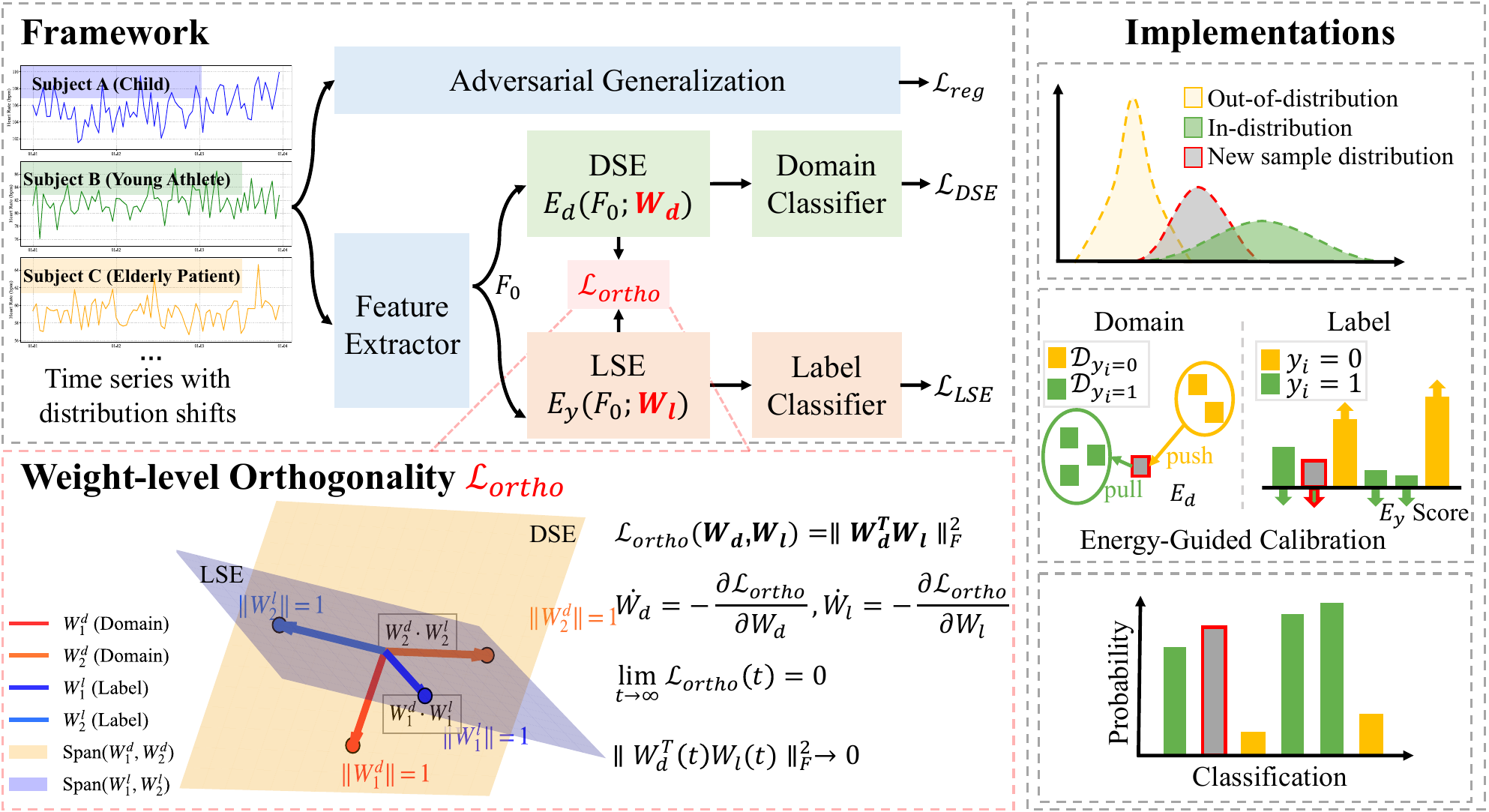} 
\caption{The \textbf{left upper} figure provides an overview of ERIS, which comprises three branches: a domain-specific energy (DSE) branch, a label-specific energy (LSE) branch, and an auxiliary adversarial generalization (AG) branch. $E_d(\cdot)$ and $E_y(\cdot)$ represent energy function. The \textbf{left lower} figure shows the weight-level orthogonality loss ($\mathcal{L}_{\mathrm{ortho}}$), which enforces asymptotic orthogonality on the spaces of the domain and label projection matrices (\textbf{\textit{Lemma}}~\ref{lem:ortho_loss}). The \textbf{right} figure illustrates the implementation of the framework. Energy calibration facilitates semantically guided feature assignment by leveraging the contrast between low energy values for true domains/labels and high energy values for others, together with prototype distances.}
\label{model}
\end{figure*}

\subsection{Energy-Guided Representation Learning}
Traditional disentanglement methods, such as those based on VAEs and GANs, commonly rely on regularization to enforce statistical independence in the latent space \cite{Xiasurvey}. However, this approach has inherent limitations. Such internal constraints do not guarantee that the learned factors will align with meaningful, real-world concepts. As a promising alternative, EBMs operate on a different principle. Instead of imposing indirect constraints, EBMs learn an energy function to directly model the probability density of the data. This direct modeling of data compatibility provides a powerful and highly flexible framework. Their broad applicability has been demonstrated in several advanced areas. For example, EOW-Softmax \cite{WangEOWSoftmax} uses EBMs for uncertainty calibration. ELI \cite{joseph2022energy} applies them in incremental learning. EGC \cite{guo2023egc} and the work by Golan et al. \cite{golan2024enhancing} show their success in image generation and classification.

In the OOD field, a well-known use of EBMs is for OOD detection. A representative work uses energy scores to identify whether a sample comes from an unseen distribution \cite{liu2020energy}. Our framework takes a fundamentally different approach. We do not treat energy as a post hoc scoring function for detection. Instead, we use it as a guiding signal that is deeply integrated into the representation learning process. In short, ERIS transforms the role of energy from a detector to a guide.

\section{Proposed Approach}

The goal of ERIS is to perform guided disentanglement of learned representations, ensuring a clear separation between domain-specific and label-relevant features. This facilitates reliable OOD generalization in TSC. The overall architecture of ERIS is depicted in Fig. \ref{model}.

\subsection{Preliminaries}
Formally, we define a time series dataset as $\mathcal{T} = \{\mathcal{X}, \mathcal{D}, \mathcal{Y}\}$, where $\mathcal{X}$, $\mathcal{D}$, and $\mathcal{Y}$ represent the collections of time series samples, domains, and labels, respectively. Each time series sample $\boldsymbol{x_i} \in \mathcal{X}$ is an ordered sequence of $m$ observations. The dataset comprises data from $N_d$ distinct domains, which may correspond to different subjects, sensors, or operating environments.

The core challenge stems from the distribution shift between the training data, $\mathcal{T}^{tr}$, and the unseen testing data, $\mathcal{T}^{te}$. The joint probability distribution of the training set differs from that of the test set, i.e., $\mathbb{P}(\mathcal{X}^{tr}, \mathcal{Y}^{tr}) \neq \mathbb{P}(\mathcal{X}^{te}, \mathcal{Y}^{te})$. Our model is trained on the multi-domain training set and must generalize to a target domain whose distribution is unknown during training.

To perform classification, a model typically employs a feature extractor, $\mathcal{F}$, to map an input time series $\boldsymbol{X}$ to an initial feature representation, $F_0 = \mathcal{F}(\boldsymbol{X})$. However, this representation is inevitably entangled, concurrently capturing both the class-discriminative patterns, which should be domain-invariant, and the domain-specific variations. The central goal of this work is to design a framework that can effectively disentangle this representation, isolating the invariant features to ensure robust generalization on unseen target domains. We summarize the primary mathematical notations in Table \ref{tab_math}.
\begin{table}[!htbp]
\setlength{\tabcolsep}{1mm}
\caption{Mathematical Notation}
\centering
\begin{tabular}{c|c}
\toprule
Notions    & Descriptions \\ \midrule
$\mathcal{T}, \mathcal{X}, \mathcal{D}, \mathcal{Y}$ & The dataset, time series, domains, and labels. \\
$\boldsymbol{x_i} \in \mathbb{R}^{N \times C}$ & An input sample with length $N$ and $C$ channels. \\
$d_i, y_i$ & The domain and label for a sample. \\
$F_0 \in \mathbb{R}^{h}$ & Shared $h$-dimensional feature representation. \\
$E_d, E_y$ & Domain-specific and label-specific energy functions. \\
$\boldsymbol{W_d}, \boldsymbol{W_l} \in \mathbb{R}^{h \times d}$ & Projection matrices for the DSE and LSE branches. \\
$P_y \in \mathbb{R}^h$ & Prototype vector for class $y$. \\
$\mathcal{L}$ & Loss function. \\
$\Theta$ & The set of all learnable model parameters. \\
\bottomrule
\end{tabular}
\label{tab_math}
\end{table}

\subsection{Energy-Guided Calibration Mechanism}

We obtain a shared representation, denoted as $F_0\in\mathbb{R}^{h}$, from a shared feature extractor. Based on this representation, the energy-guided calibration mechanism employs two strategies:  an energy-based semantic guidance and a weight-level orthogonality loss (see bottom left of Fig. \ref{model}). The former is achieved by defining two types of learnable energy functions: domain-specific energy ($E_d$) and label-specific energy ($E_y$). These are parameterized functions that take the shared feature representation $F_0$ as input and map it to a scalar energy value. The parameters of these functions are learned to instill semantic guidance, where a lower energy value signifies a higher compatibility between the input features and the corresponding domain or label.  Note that semantic guidance here refers to learned discriminative signals, not inherent properties of the data. We denote these functions as:
\begin{equation}
\begin{aligned}
E_d&= \mathcal{E}_d(F_0; \boldsymbol{\theta_d}), && \forall d \in \mathcal{D}, \\
E_y&= \mathcal{E}_y(F_0; \boldsymbol{\theta_y}), && \forall y \in \mathcal{Y},
\end{aligned}
\end{equation}
where $\mathcal{E}_d(\cdot)$ and $\mathcal{E}_y(\cdot)$ represent the energy functions parameterized by learnable weights $ \boldsymbol{\theta_d}$ and $\boldsymbol{\theta_y}$, respectively.
Lower energy values $E_d$ and $E_y$ indicate better alignment with the target domain and label semantics, respectively. For both values, lower is better. We guide the function to quantify semantic guidance rooted in physical properties, such as the inherent differences in signal variance that exist across domains. To further enforce the separation, a weight-level orthogonality loss is applied to the projection matrices ($\boldsymbol{W_d}, \boldsymbol{W_l}$). This constraint ensures the domain and label subspaces are structurally independent, promoting a clean disentanglement.

\begin{figure}
\centering
\includegraphics[width=0.95\columnwidth]{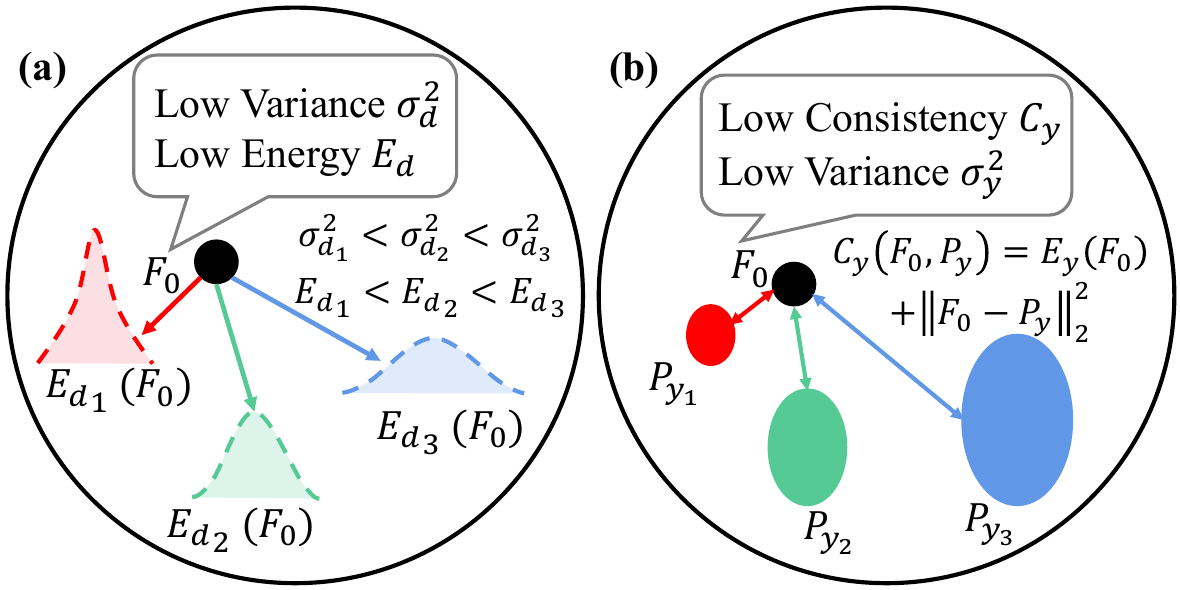} 
\caption{Conceptual illustration of domain-specific energy (DSE) and label-specific energy (LSE). (a) The left figure illustrates the concept of DSE, in which lower energy signifies a more concentrated data distribution.
(b) The right figure depicts LSE, a metric where lower energy reflects higher consistency between a feature and its corresponding label.}
\label{fig_energy}
\end{figure}

\textbf{Domain-Specific Energy (DSE).} The objective is to assign lower energy to features originating from the true domain $d_{\text{true}}$, and higher energy to features under alternative domains. The DSE energy functions serve as indicators of how well a feature generalizes across different domains. Lower energy values indicate better generalization. To optimize these energy functions, DSE employs a contrastive structured loss defined as:
\begin{equation}\label{eq:dse}
\begin{aligned}
\mathcal{L}_{\text{DSE}} =
\mathbb{E}_{(x, d) \sim \mathcal{D}_S} \Big[
E_{d} +
\sum_{d' \neq d}
\max(0, m_\mathcal{D} - E_{d'})
\Big],
\end{aligned}
\end{equation}
where $m_\mathcal{D}$ is an energy margin. This contrastive formulation encourages minimization of the energy for the true domain while pushing the energies associated with other domains above the margin, thereby enhancing the model's calibration by learning to distinguish domain-specific features. As conceptually illustrated in Fig. \ref{fig_energy} (a), this process associates lower domain-specific energy with a more concentrated data distribution (i.e., lower variance).
Moreover, the DSE also models the domain-specific predictive uncertainty $\sigma_d^2$ and enforces that the confidence scores align with the negative energy values, which serve as proxies for domain match quality. This constraint can be formalized as:
\begin{align}
\arg \max_{\{\sigma_d^2\}} \ 
\text{Correlation} \left(
\text{Rank}(\|\sigma_d^2\|_2^{-1}), \
\text{Rank}(-E_d)
\right),
\end{align}
where the operations are applied over all domains $d \in \mathcal{D}$, $\|\sigma_d^2\|_2$ denotes the norm of the variance vector, and $-E_d$ reflects the degree of domain match—the higher the value, the better the match. Although the DSE training loss does not directly optimize the ranking objective above, minimizing $\mathcal{L}_{\text{DSE}}$ implicitly facilitates the learning of such ordinal relations in practice. 

\textbf{Label-Specific Energy with Prototype Congruence (LSE).} The LSE is designed to enhance the model's discriminative capability. Similar to DSE, LSE defines a set of label-specific energy functions, and incorporates class prototypes $P_y$ to reinforce the consistency between features and their corresponding labels. It should be assigned a lower consistency energy if the input feature $F_0$ belongs to the ground-truth label $y_{\text{true}}$ and lies close to its prototype $P_{y_{\text{true}}}$. To jointly quantify both label-specific energy and prototype distance, LSE can be defined as a consistency error function that 
\begin{equation}
C_y= E_y + \|F_0 -P_y\|_2^2.
\end{equation}
This function reflects how well the feature $F_0$ aligns with class $y$: smaller values indicate higher compatibility due to lower energy and closer proximity to the prototype. This function, conceptually visualized in Fig. \ref{fig_energy} (b), reflects how well the feature $F_0$ aligns with class y: smaller values indicate higher compatibility due to lower energy and closer proximity to the prototype.

The LSE loss comprises two components. The first is the label-wise contrastive loss:
\begin{equation}\label{eq:cl}
\begin{aligned}
\mathcal{L}_{\text{CL}} =
\mathbb{E}_{(x, y) \sim \mathcal{D}_S} \Big[
E_{y} +
\sum_{y' \neq y}
\max\left(0, m_{L_1} - E_{y'} \right)
\Big].
\end{aligned}
\end{equation}
This objective encourages minimization of energy for the actual label while pushing incorrect label energies above a margin $m_{L_1}$. The second component is the prototype contrastive loss:
\begin{equation}\label{eq:proto}
\begin{aligned}
\mathcal{L}_{\text{Proto}} =\;&
\mathbb{E}_{(x, y) \sim \mathcal{D}_S} \Big[
\|F_0 - P_y\|_2^2 \\
&\quad - \frac{1}{N_C - 1} \sum_{y' \neq y}
\left( \|F_0 - P_{y'} \|_2^2 + m_{L_2} \right)
\Big].
\end{aligned}
\end{equation}
This loss encourages features to align closely with the actual class prototype and be distant from other class prototypes. The margin $m_{L_2}$ controls the strength of repulsion.

The final LSE loss is given by:
\begin{equation}\label{eq:lse_total}
\mathcal{L}_{\text{LSE}} = \mathcal{L}_{\text{CL}} + \mathcal{L}_{\text{Proto}}.
\end{equation}
Through the minimization of $\mathcal{L}_{\text{LSE}}$, a link is established between low consistency error and high confidence, allowing for a structured representation of label-level fit quality. The resulting metric $C_y$ is regarded as a calibrated measure of label compatibility, which serves to guide discriminative representation learning.

\textbf{Weight-level Orthogonality Loss.} 
In constructing the objective function, the original domain and label energy loss, are formulated using contrastive learning or energy minimization. When combined, these losses drive the two respective branches to focus on their tasks. However, this alone does not guarantee that the subspaces learned by the two networks are mutually orthogonal. Their parameter spaces may still overlap, leading to interference between domain and label information. To address this, we introduce an orthogonality penalty term:
\begin{align}\label{eq:ortho}
    \mathcal{L}_{\mathrm{ortho}}(\boldsymbol{W_d},\boldsymbol{W_l})=\bigl\lVert \boldsymbol{W_d^{\mathsf T}W_l}\bigr\rVert_F^{2}= {\textstyle \sum_{ij}} (\boldsymbol{W_d,W_l})_{i,j}^{2}.
\end{align}
Unlike feature-level orthogonality \cite{qian2021,shi2024}, our framework enforces weight-level orthogonality. The key distinction lies in the scope of the constraint. Feature-level orthogonality is a per-instance constraint, enforced on the output features of each training sample. This can lead to brittle solutions that are sensitive to data shifts, as the underlying projection functions may remain correlated. In contrast, our weight-level orthogonality imposes a global structural guidance by directly making the model's projection matrices ($\boldsymbol{W_d}$,$\boldsymbol{W_l}$) orthogonal. This ensures the projection functions themselves are decorrelated, providing a more robust guarantee of feature separation for any input, including OOD samples. This term reaches its global minimum when $\boldsymbol{W_d^{\mathsf T}W_l}=0$ and applies a quadratic penalty to any non-zero entries in the matrix. During optimization via gradient descent or its variants, this orthogonality penalty continuously pushes the updates in a direction that forces $\boldsymbol{W_d^T W_l}$ tends to 0. This ensures that the column spaces of the two projection matrices, $\boldsymbol{W_d}$ and $\boldsymbol{W_l}$, are driven towards orthogonality.

\textit{\textbf{Lemma:}} Let \(\boldsymbol{W_d},\boldsymbol{W_l}\in\mathbb{R}^{h\times d}\) and define the weight-level orthogonality loss:
\begin{align}\label{eq:ortho}
    \mathcal{L}_{\mathrm{ortho}}(\boldsymbol{W_d},\boldsymbol{W_l})=\bigl\lVert \boldsymbol{W_d^{\mathsf T}W_l}\bigr\rVert_F^{2}.
\end{align}
As shown in the lower-left part of Fig. 2, the weight-level orthogonality loss decouples the parameter spaces of $W_d$ and $W_l$ by penalizing their non-zero inner product. 
Then \(\mathcal{L}_{\mathrm{ortho}}(t)\) is non-increasing along every trajectory and $ \lim_{t\to\infty} \mathcal{L}_{\mathrm{ortho}}(t)=0.$
Consequently \(\lVert \boldsymbol{W_d^{\mathsf T}(t) W_l(t)}\rVert_F \) tends to 0; in particular every non-degenerate \(\omega\)-limit point $(\boldsymbol{W_d^*},\boldsymbol{W_l^*})$ satisfies  the condition that $(\boldsymbol{W_d^*},\boldsymbol{W_l^*})=0$.
\label{lem:ortho_loss}

\begin{IEEEproof}[\textbf{Proof}]
Write the weight-level orthogonality loss in trace form,
\begin{equation}
\begin{split}
    \mathcal{L}_{\mathrm{ortho}}
      & =\operatorname{Tr}\!\bigl[(\boldsymbol{W_d^{\mathsf T}W_l)^{\mathsf T}(W_d^{\mathsf T}W_l)}\bigr] \\
      & =\operatorname{Tr}\!\bigl(\boldsymbol{W_l^{\mathsf T}W_d\,W_d^{\mathsf T}W_l}\bigr).
\end{split}
\end{equation}
Standard matrix calculus yields
$
    \nabla_{W_d}\mathcal{L}_{\mathrm{ortho}} = 2\boldsymbol{W_lW_l^{\mathsf T}W_d}
$
and $\nabla_{W_l}\mathcal{L}_{\mathrm{ortho}} = 2\boldsymbol{W_dW_d^{\mathsf T}W_l}.$
Hence, the gradient flow can be written explicitly as:
\begin{align}
    \dot W_d=-2\,\boldsymbol{W_lW_l^{\mathsf T}W_d},\qquad
    \dot W_l=-2\,\boldsymbol{W_dW_d^{\mathsf T}W_l}.
\end{align}
The Frobenius inner product \(\langle \boldsymbol{A},\boldsymbol{B}\rangle=\operatorname{Tr}(\boldsymbol{A^{\mathsf T}B})\) gives
\begin{align*}
 \frac{\mathrm d}{\mathrm dt}\mathcal{L}_{\mathrm{ortho}}
&= -4 \left\| \boldsymbol{W_l W_l^\top W_d} \right\|_F^2 
- 4 \left\| \boldsymbol{W_d W_d^\top W_l}\right\|_F^2 
\leq 0.
\end{align*}
So $\mathcal{L}_{\mathrm{ortho}}(t)$ is monotone decreasing and bounded below by 0; the limit $\mathcal{L}_\infty:=\lim_{t \to \infty} \mathcal{L}_{\mathrm{ortho}}(t)$ therefore exists.
Assume for contradiction that $\mathcal{L}_\infty$ is greater than 0. Convergence of $\mathcal{L}_{\mathrm{ortho}}(t)$ implies $\mathcal{\dot L} _{\mathrm{ortho}}(t)$ that it tends to 0, which forces
$
     \lVert \boldsymbol{\boldsymbol{W_lW_l^{\mathsf T}W_d}}\rVert_F^2
$ and
$
     \lVert \boldsymbol{\boldsymbol{W_dW_d^{\mathsf T}W_l}}\rVert_F^2
$ to grow to 0. 
Choose a sequence $t_k$ tends to infinite and set \(\boldsymbol{A_k}:=\boldsymbol{W_l(t_k)^{\mathsf T}W_d(t_k)}\). Because $ 
    \lVert \boldsymbol{W_l(t_k)A_k}\rVert_F^2
$ tends to 0
and \(\boldsymbol{W_l^{\mathsf T}W_l}\) is positive semidefinite, multiplying on the left by \(\boldsymbol{W_l(t_k)^{\mathsf T}}\) and using the pseudo-inverse on its support yields \(\lVert \boldsymbol{A_k}\rVert_F^{2}\) tends to 0. Hence, \(\mathcal{L}_{\mathrm{ortho}}(t_k)=\lVert \boldsymbol{A_k}\rVert_F^{2}\) tends to 0, contradicting $\mathcal{L}_\infty$ is greater than 0. Thus, \(\mathcal{L}_\infty\) equals 0, and the claimed limit follows. Any \(\omega\)-limit point must satisfy \(\boldsymbol{W_l^{\mathsf T}W_d}=0\) because \(\mathcal{L}_{\mathrm{ortho}}\) is continuous.
\end{IEEEproof}
With the minimization of the orthogonality loss, the feature representations produced by the domain (DSE) and label (LSE) branches are gradually driven toward asymptotic orthogonality in the latent space. Consequently, label information is effectively separated from domain-specific variations.

\subsection{Adversarial Generalization (AG)}

We propose adversarial generalization (AG), a mechanism designed to foster robust and generalizable representations through a two-fold adversarial strategy. Firstly, to achieve global alignment, it employs an adversarial game where a feature extractor is trained to deceive a domain discriminator. This forces the learned features to become domain-indistinguishable, thereby aligning distributions across different domains. Secondly, to enhance local robustness, AG smooths the latent feature space. It identifies an adversarial perturbation $r_{\text{adv}}$ that maximally alters the model's predictive distribution: 
\begin{equation}\label{eq:adv}
r_{\text{adv}} = \arg \max_{\|r\|_2 < \epsilon} 
D_{\mathrm{KL}} \left( p(y \mid h, \hat{\theta}) \ \| \ p(y \mid h + r, \hat{\theta}) \right).
\end{equation}
Subsequently, the model is trained to be resilient against this perturbation by minimizing the regularization loss:
\begin{equation}\label{eq:reg}
\mathcal{L}_{\text{reg}} = D_{\mathrm{KL}} \left( 
p(y \mid h, \hat{\theta}) \ \| \ p(y \mid h + r_{\text{adv}}, \theta) \right).
\end{equation}
By synergizing global domain alignment with local adversarial smoothing, AG produces representations that are both domain-agnostic and locally invariant, leading to superior generalization under distribution shifts.

\subsection{Model Summary}

\begin{algorithm}[t]
\label{algo}
\caption{Training Pipeline of ERIS}
\KwIn{Time series training set $\mathcal{T}^{\mathrm{tr}}\!=\!\{\mathcal{X}^{\mathrm{tr}},\mathcal{D} ^{\mathrm{tr}},\mathcal{Y} ^{\mathrm{tr}}\}$; learning rate $\eta$; perturbation budget $\epsilon$.}
\BlankLine
Initialize model parameters $\Theta$;\\
\While{not converged}{
Sample batch $(x,d,y)$ from $\mathcal{T}^{\mathrm{tr}}$;
$F_{0} \leftarrow \textsc{FeatureExtractor}(x)$ \tcp*{Shared feature extraction}
$\mathcal{L}_{\text{DSE}}\leftarrow \textsc{Energy}(\mathcal{E}_{d}(F_{0};\boldsymbol{\theta_d}), d)$ \tcp*{Domain energy (Eq.\ref{eq:dse})}
$\mathcal{L}_{\text{LSE}} \leftarrow \textsc{Energy}(\mathcal{E}_{y}(F_{0};\boldsymbol{\theta_y}), y)$ \tcp{Label energy (Eq.\ref{eq:cl} to Eq.\ref{eq:lse_total})}
$\mathcal{L}_{\text{orth}} \leftarrow |\mathbf{W}_d^{\top}\mathbf{W}_l\|_{F}^{2}$ \tcp{Weight orthogonality (Eq.\ref{eq:ortho})}
$\mathcal{L}_{\text{reg}} \leftarrow \textsc{AdvRegularization}(h, y, \theta)$ \tcp*{Adversarial regularization (Eq.\ref{eq:adv} and Eq.\ref{eq:reg})}
$\mathcal{L}{\text{total}} \leftarrow \lambda_{1}\mathcal{L}_{\text{DSE}} + \lambda_{2}\mathcal{L}_{\text{LSE}} + \mathcal{L}_{\text{orth}} + \mathcal{L}_{\text{reg}}$ \tcp*{Total loss (Eq.\ref{eq:total})}
Update $\Theta$ with $\nabla_{\Theta}\mathcal{L}_{\mathsf{total}}$;
}
\KwOut{Optimized parameters $\Theta^{\star}$.}
\end{algorithm}

In summary, we adopt an end-to-end training strategy where the total loss function is defined as a weighted sum of the individual objectives:
\begin{equation}\label{eq:total}
\mathcal{L}_{\text{total}} = \lambda_1 \mathcal{L}_{\text{DSE}} + \lambda_2 \mathcal{L}_{\text{LSE}} + \mathcal{L}_{\text{ortho}} + \mathcal{L}_{\text{reg}},
\end{equation}
where $\lambda_1$ and $\lambda_2$ are the trade-off parameters between the loss functions. Empirically, we set a relatively small $\lambda_1$ and a larger $\lambda_2$, so that the prototype-based LSE loss dominates the learning process to ensure strong classification performance under unseen or OOD settings. During inference, no additional computation of the auxiliary and orthogonality loss terms is required, significantly simplifying the deployment phase. The training pipeline of ERIS is described in \textbf{Algorithm} \ref{algo}.

Let the input time series length be $N$, and the per-timestep channel dimension be $d_{\text{in}}$. The input is first processed by $L$ layers of 1D convolution, where each layer uses kernel size $K$, and outputs $C_l$ channels. The resulting feature maps are then aggregated via global pooling (or flattening), producing an encoding vector $x \in \mathbb{R}^{1 \times b}$. Two separate 3-layer MLPs, each with a hidden width of $h$, are used for the label and domain branches, respectively. Their output dimensions correspond to the number of labels, $N_y$, and the number of domains, $N_d$.

\textbf{Time.} The convolutional encoding phase consumes $O(LNKC^2)$ under uniform configurations. Next, the two 3-layer MLPs perform forward propagation, incurring a combined cost of
\begin{equation}
O\left(bh + h^2 + h(N_y + N_d)\right).
\end{equation}
Combining both parts and dropping lower-order constants yields an overall forward-pass complexity of:
\begin{equation}
O\left(LNKC^2 + bh + h^2 + h(N_y + N_d)\right).
\end{equation}
As the output layer cost $h(N_y + N_d)$ is typically dominated by $h^2$, we simplify the expression to:
\begin{equation}
O\left(LNKC^2 + h(b + h)\right).
\end{equation}

\textbf{Space.} Model parameters consist of the convolution kernels, requiring $O(LNKC^2)$, and the MLP weights, contributing 
\begin{equation}
O(bh + h^2 + h(N_y + N_d)). 
\end{equation}
For dynamic memory, the model needs to store the input sequence $O(N)$, the encoded feature $O(b)$, and the output tensors $O(2N_y + 2N_d)$. As with time complexity, $h(N_y + N_d)$ is typically absorbed into $h^2$, resulting in the final space complexity:
\begin{equation}
O\left(LNKC^2 + h(b + h) + N + b\right).
\end{equation}

\section{Experiments}

In this section, we conduct a series of comprehensive experiments to rigorously evaluate the ERIS framework. Our evaluation begins in Section \ref{ex_setting}, where we detail the experimental settings. Following this, Section \ref{ex_main} presents the main results. To conclude, Section \ref{sec:ab} provides in-depth ablation studies and visual analyses to break down the contributions of ERIS's key components and showcase its feature disentanglement capabilities.

\subsection{Experimental Settings}
\label{ex_setting}

\subsubsection{Datasets} We perform experiments on four large sensor benchmark datasets, each containing multiple domains and labels. The four datasets used in the experiments are UCIHAR \cite{anguita2012}, UniMiB-SHAR \cite{hnoohom2017}, Opportunity \cite{chavarriaga2013}, and EMG \cite{lobov2018}, providing a comprehensive assessment of ERIS’s OOD generalization performance. Specifically, \textbf{UCIHAR} comprises smartphone sensor data, including accelerometer and gyroscope readings, for the classification of fundamental human activities. \textbf{UniMiB-SHAR} encompasses smartphone accelerometer data for fine-grained recognition, distinguishing between daily activities and various types of falls. \textbf{EMG} is composed of electromyographic signals for hand gesture classification; its inherent sensitivity to environmental and physiological factors renders it a particularly challenging benchmark for OOD generalization. Finally, \textbf{Opportunity} is a richly annotated multimodal dataset aggregated from a diverse array of wearable and environmental sensors, intended for the task of recognizing complex activities of daily living within a residential setting. The statistics of these datasets is summarized in Table \ref{tab_data}

\begin{table}[!htbp]
\setlength{\tabcolsep}{2mm}
\caption{Summary of four datasets.}
\centering
\begin{tabular}{ccccc}
\toprule
Datasets    & Samples  & Channels & Classes & Domains \\ \midrule
UCIHAR      & 1,609   & 9       & 6     & 5      \\
UniMiB-SHAR & 1,569   & 453     & 17    & 4      \\
Opportunity & 869,387 & 77      & 18    & 4      \\
EMG         & 6,883   & 200     & 6     & 4    \\  \bottomrule
\end{tabular}
\label{tab_data}
\end{table}

\subsubsection{Baselines} To demonstrate the effectiveness of our proposed ERIS model, we compare it against a comprehensive set of baselines, which are grouped into four distinct categories:
\begin{itemize}
\item Energy-based methods. We include several established energy-based models in our comparison, including EOW Softmax \cite{WangEOWSoftmax}, ELI \cite{joseph2022energy}, EGC \cite{guo2023egc}, and Golan \cite{golan2024enhancing}.
\item General time series methods. Our evaluation includes representative general time series methods, namely TARNet \cite{chowdhury2022tarnet}, PatchTST \cite{nie2023a}, TS-GAC \cite{wang2024graph}, and MPTSNet \cite{mu2025mptsnet}.
\item General OOD generalization methods. We select prominent OOD generalization baselines from broader research domains, such as GroupDRO \cite{Sagawa2020Distributionally}, VREx \cite{krueger2021}, SDL \cite{ye2023}, FOOD \cite{liao2024}, and EVIL \cite{huang2025}.
\item Time series OOD generalization methods. This category comprises four significant works specifically designed for time series OOD generalization, including GILE \cite{qian2021}, AdaRNN \cite{du2021}, Diversify \cite{lu2023}, and ITSR \cite{shi2024}.
\end{itemize}

\begin{table*}
\caption{Overall comparison results of ERIS and other methods across four datasets (Acc, in \%, $\uparrow$). 'Avg.' is short for average. 'Avg. Rank is determined by summing the ranks in each dataset and dividing the total by the number of datasets considered. 'Avg. ALL' represents the average accuracy across all datasets. The $p$-value is obtained using the Wilcoxon signed-rank test; a value less than 0.05 indicates that the performance difference between ERIS and the compared method is statistically significant. Top results: \textbf{\textcolor{darkred}{1st}}, \textbf{\textcolor{darkblue}{2nd}}.}
\setlength{\tabcolsep}{1.8mm} 
\begin{tabular}{cc|ccccc|cccc}
\toprule
\multirow{2}{*}{Datasets}                                               & \multirow{2}{*}{\begin{tabular}[c]{@{}c@{}}Target\\ Domain\end{tabular}}                      & \multirow{2}{*}{\begin{tabular}[c]{@{}c@{}}ERIS\\ (Ours)\end{tabular}} & \multicolumn{4}{|c|}{Energy-based}                                                                                                                                                                                                                                                      & \multicolumn{4}{c}{Time Series OOD Generalization}                                                                                                                                                                                                         \\ \cmidrule{4-11} 
&
                                                                        &                                                                                               &                                                                        \multicolumn{1}{|c}{\begin{tabular}[c]{@{}c@{}}EOW-Softmax\\ ICCV'21\end{tabular}} & \begin{tabular}[c]{@{}c@{}}ELI\\ CVPR'22\end{tabular}  & \multicolumn{1}{c}{\begin{tabular}[c]{@{}c@{}}EGC\\ ICCV'23\end{tabular}} & \multicolumn{1}{c|}{\begin{tabular}[c]{@{}c@{}}Golan\\ NeurIPS'24\end{tabular}} & \begin{tabular}[c]{@{}c@{}}GILE\\ AAAI '21\end{tabular} & \begin{tabular}[c]{@{}c@{}}AdaRNN\\ CIKM'21\end{tabular} & \begin{tabular}[c]{@{}c@{}}Diversify\\ ICLR'23\end{tabular}    & \begin{tabular}[c]{@{}c@{}}ITSR\\ KDD'24\end{tabular} \\ \midrule
\multirow{7}{*}{UCIHAR}                                                 & 0                                                                                             &\multicolumn{1}{c|}{\textbf{\textcolor{darkred}{94.70\tiny{±1.20}}}}                &94.15\tiny{±0.60}	 &91.07\tiny{±0.65}&87.36\tiny{±1.46}	&85.45\tiny{±1.73}                   & 79.38\tiny{±1.63}                                                & 82.83\tiny{±1.64}                                                   & 79.25\tiny{±5.01}                                                 & 78.80\tiny{±5.58}                                                    \\
                                                                        & 1                                                                                                                                                               & \multicolumn{1}{c|}{\textbf{\textcolor{darkred}{84.37\tiny{±2.17}}}}           & \textbf{\textcolor{darkblue}{84.14\tiny{±0.60}}}	 & 79.48\tiny{±1.00}	& 73.04\tiny{±3.86} & 74.27\tiny{±2.74}          & 75.63\tiny{±0.95}                                                & 60.80\tiny{±5.73}                                                   & 55.77\tiny{±6.40}                                                 & 80.67\tiny{±4.65}                                                      \\
                                                                        & 2                                                                                             &                                                                        \multicolumn{1}{c|}{\textbf{\textcolor{darkred}{95.05\tiny{±1.00}}}}             &\textbf{\textcolor{darkblue}{94.55\tiny{±0.31}}}& 80.53\tiny{±0.74}& 85.48\tiny{±3.03}& 85.18\tiny{±3.47}       & 91.91\tiny{±2.34}                                                & 78.13\tiny{±4.61}                                                   & 93.61\tiny{±2.44}                                                & 91.97\tiny{±0.99}                          \\
                                                                        & 3                                                                                             &                                                                        \multicolumn{1}{c|}{\textbf{\textcolor{darkred}{94.39\tiny{±2.84}}}}             &  \textbf{\textcolor{darkblue}{93.99\tiny{±0.80}}}	& 86.00\tiny{±2.31}	& 83.04\tiny{±3.14} &85.40\tiny{±1.44}                    & 91.13\tiny{±1.86}                                                & 69.85\tiny{±4.23}                                                  & 66.44\tiny{±4.45}                                                 & 86.06\tiny{±2.73}                                                                             \\
                                                                        & 4                                                                                             &                                                                        \multicolumn{1}{c|}{\textbf{\textcolor{darkred}{97.42\tiny{±0.98}}}}             & 95.70\tiny{±1.02}	& 84.31\tiny{±0.65} & 84.96\tiny{±3.59} &	84.00\tiny{±2.99}               & 87.25\tiny{±2.11}                                                & 81.86\tiny{±3.43}                                                   & 73.91\tiny{±3.16}                                                 & 91.33\tiny{±2.88}                               \\  
                                                                        &\cellcolor{gray!12} Avg. Acc                                                                                      &    \multicolumn{1}{c|}{\cellcolor{gray!12}\textbf{\textcolor{darkred}{93.19\tiny{±4.91}}}}          &\cellcolor{gray!12}\textbf{\textcolor{darkblue}{92.51\tiny{±4.36}}}&\cellcolor{gray!12}84.28\tiny{±4.39} & \cellcolor{gray!12}82.78\tiny{±5.90}	&\cellcolor{gray!12}82.86\tiny{±5.01}                            & \cellcolor{gray!12}83.82\tiny{±5.87}                                               & \cellcolor{gray!12}77.45\tiny{±11.63}                                                   & \cellcolor{gray!12}70.70\tiny{±9.93}                                                 & \cellcolor{gray!12}86.09\tiny{±6.87}                            \\ 
                                                                        & \cellcolor{gray!12} Avg. Rank                                                                                     &  \multicolumn{1}{c|}{\cellcolor{gray!12}\textbf{\textcolor{darkred}{1.00}}}&                           \cellcolor{gray!12}\textbf{\textcolor{darkblue}{2.60}}                                 &  \cellcolor{gray!12}11.00                                                      &      \cellcolor{gray!12}12.60                                                                      &                \cellcolor{gray!12} 12.60                                                                &         \cellcolor{gray!12}10.20                                                                                                      &            \cellcolor{gray!12}15.20                                                &              \cellcolor{gray!12}17.60 &\cellcolor{gray!12} 9.00                                             \\ \midrule
\multirow{6}{*}{\begin{tabular}[c]{@{}c@{}}UniMiB-\\ SHAR\end{tabular}} & 1                                                                                             &        \multicolumn{1}{c|}{\textbf{\textcolor{darkred}{59.59\tiny{±0.95}}}}       &\textbf{\textcolor{darkblue}{57.07\tiny{±3.52}}} &35.63\tiny{±0.29} &19.72\tiny{±14.89}&32.46\tiny{±12.76}               & 46.93\tiny{±1.55}                                                & 34.95\tiny{±5.66}                                                   & 50.94\tiny{±3.21}                                                 & 51.67\tiny{±2.67}                                                              \\ 
                                                                        & 2                                                                                             &                                                                         \multicolumn{1}{c|}{\textbf{\textcolor{darkblue}{49.34\tiny{±0.83}}}}      &43.85\tiny{±2.66} &	28.27\tiny{±1.95} &	20.60\tiny{±13.59} &	32.00\tiny{±3.75}                   & 47.97\tiny{±2.26}                                                & 36.23\tiny{±5.58}                                                   & 46.04\tiny{±3.25}                                                 & \textbf{\textcolor{darkred}{50.60\tiny{±4.42}}}                  \\
                                                                        & 3                                                                                             &                                                                        \multicolumn{1}{c|}{\textbf{\textcolor{darkred}{70.60\tiny{±1.59}}}}      & \textbf{\textcolor{darkblue}{70.13\tiny{±2.14}}}	&40.07\tiny{±2.02}	&19.48\tiny{±16.47}	&28.79\tiny{±15.73}             & 66.82\tiny{±1.62}                                                & 39.41\tiny{±8.17}                                                   & 57.70\tiny{±2.74}                                                 & 58.95\tiny{±7.37}                        \\
                                                                        & 5                                                                                             &                                                                        \multicolumn{1}{c|}{41.41\tiny{±0.81}}             &41.31\tiny{±0.86}	&34.43\tiny{±1.97}&	34.20\tiny{±2.50}&	34.81\tiny{±3.25}                             & 37.38\tiny{±1.27}                                                & 33.96\tiny{±7.73}                                                   & 41.68\tiny{±1.89}                                                 & \textbf{\textcolor{darkblue}{43.15\tiny{±2.14}}}                                          \\
                                                                        & \cellcolor{gray!12}Avg. Acc                                                                                      &   \multicolumn{1}{c|}{\cellcolor{gray!12} \textbf{\textcolor{darkred}{55.23\tiny{±11.30}}}}      &\cellcolor{gray!12}\textbf{\textcolor{darkblue}{53.09\tiny{±12.04}}}	&\cellcolor{gray!12}34.60\tiny{±4.61}	&\cellcolor{gray!12} 23.50\tiny{±13.58}	&\cellcolor{gray!12}32.02\tiny{±9.82}                         & \cellcolor{gray!12}49.77\tiny{±11.06}                                                & \cellcolor{gray!12}36.14\tiny{±6.66}                                                  & \cellcolor{gray!12}49.09\tiny{±6.64}                                          & \cellcolor{gray!12}51.10\tiny{±7.15}                                                                                \\
  & \cellcolor{gray!12}Avg. Rank                                                                                     & \multicolumn{1}{c|}{ \cellcolor{gray!12}\textbf{\textcolor{darkred}{2.00}}}	&\cellcolor{gray!12}4.25	&	\cellcolor{gray!12}15.50		&\cellcolor{gray!12}17.50	&	\cellcolor{gray!12}16.00	&\cellcolor{gray!12}7.00	&	\cellcolor{gray!12} 15.75	&	\cellcolor{gray!12}6.75	&	\cellcolor{gray!12}4.75                              \\ \midrule
\multirow{6}{*}{EMG}                                                    & 0                                                                                             &                                                                        \multicolumn{1}{c|}{\textbf{\textcolor{darkred}{73.59\tiny{±1.84}}}}        & 65.03\tiny{±4.68}&52.27\tiny{±3.71}&54.44\tiny{±2.27}	&49.83\tiny{±2.78}             & 39.27\tiny{±4.19}    & 52.09\tiny{±3.18}                                                   & 48.99\tiny{±4.97}                                                 & \textbf{\textcolor{darkblue}{67.17\tiny{±1.99}}}             \\
                                                                        & 1                                                                                             &                                                                        \multicolumn{1}{c|}{\textbf{\textcolor{darkred}{49.11\tiny{±2.58}}}}           &44.60\tiny{±2.26}	&34.98\tiny{±1.55}	&38.96\tiny{±2.54}&	48.74\tiny{±3.60}               & 38.57\tiny{±6.12}             & 44.09\tiny{±1.36}              & 42.52\tiny{±6.66}              & \textbf{\textcolor{darkblue}{48.80\tiny{±3.71}}}        \\
                                                                        & 2                                                                                             &                                                                        \multicolumn{1}{c|}{\textbf{\textcolor{darkred}{65.12\tiny{±6.20}}}}       &\textbf{\textcolor{darkblue}{64.40\tiny{±4.58}}}&	48.96\tiny{±1.12}&	36.96\tiny{±14.38}	&61.87\tiny{±6.07}                   & 40.45\tiny{±1.99}                                                & 57.68\tiny{±2.81}                                                   & 45.12\tiny{±5.25}                                                 & 62.21\tiny{±2.06}     \\
                                                                        & 3                                                                                             &                                                                        \multicolumn{1}{c|}{\textbf{\textcolor{darkred}{85.87\tiny{±1.55}}}}        &78.66\tiny{±0.74}&	45.10\tiny{±1.20}	&40.64\tiny{±15.45}&	68.08\tiny{±4.43}                     & 50.85\tiny{±1.81}                                                & 54.62\tiny{±3.85}                                                   & 63.85\tiny{±6.52}                                                 & \textbf{\textcolor{darkblue}{82.89\tiny{±1.26}}}                                     \\
                                                                        & \cellcolor{gray!12}Avg. Acc                                                                                                                                                         &   \multicolumn{1}{c|}{\cellcolor{gray!12} \textbf{\textcolor{darkred}{68.42\tiny{±14.11}}}}      &\cellcolor{gray!12}63.17\tiny{±12.86}&	\cellcolor{gray!12}45.33\tiny{±6.96}&	\cellcolor{gray!12}42.75\tiny{±12.08}&	\cellcolor{gray!12}57.13\tiny{±9.29}                     & \cellcolor{gray!12}42.28\tiny{±6.27}                   & \cellcolor{gray!12}52.12\tiny{±5.84}                                                   & \cellcolor{gray!12}50.12\tiny{±10.05}     & \cellcolor{gray!12}\textbf{\textcolor{darkblue}{65.27\tiny{±12.71}}}                           \\
 &\cellcolor{gray!12} Avg. Rank                                                                                     & \multicolumn{1}{c|}{\cellcolor{gray!12}\textbf{\textcolor{darkred}{1.00}}}	&\cellcolor{gray!12}3.50	&	\cellcolor{gray!12}12.75	&	\cellcolor{gray!12}14.00		&\cellcolor{gray!12} 5.50&\cellcolor{gray!12}15.00&		\cellcolor{gray!12}7.75	&	\cellcolor{gray!12}9.50	&	\cellcolor{gray!12}\textbf{\textcolor{darkblue}{2.25}}                    \\ \midrule
\multirow{6}{*}{\begin{tabular}[c]{@{}c@{}}Opportu\\ nity\end{tabular}} & S1                                                                                            &\multicolumn{1}{c|}{82.22\tiny{±0.16}}           &82.02\tiny{±0.93}	&81.69\tiny{±0.51}&	74.20\tiny{±3.86}	&78.40\tiny{±2.41}                  & \textbf{\textcolor{darkblue}{83.06\tiny{±0.42}}}                                                & 80.59\tiny{±0.62}                                          & 79.37\tiny{±0.35}                                                & \textbf{\textcolor{darkred}{83.39\tiny{±0.25}}}                                                                 \\
                                                                        & S2                                                                                            &                                                                        \multicolumn{1}{c|}{\textbf{\textcolor{darkred}{81.10\tiny{±0.28}}}}     &80.59\tiny{±0.98}	&80.85\tiny{±0.12}	&74.73\tiny{±3.08}&	78.00\tiny{±2.00}             & \textbf{\textcolor{darkblue}{81.05\tiny{±0.23}}}                                                & 80.27\tiny{±0.44}                                                   & 77.80\tiny{±0.75}                                                 & 79.84\tiny{±0.45}                \\
                                                                        & S3                                                                                            &                                                                        \multicolumn{1}{c|}{\textbf{\textcolor{darkred}{77.65\tiny{±0.11}}}}      &76.81\tiny{±0.60}&	75.15\tiny{±0.78}&72.85\tiny{±3.06}&	77.60\tiny{±1.82}           & \textbf{\textcolor{darkblue}{77.62\tiny{±0.50}}}            & 74.31\tiny{±0.52}               & 73.68\tiny{±1.29}          & 76.19\tiny{±0.57}             \\
 & S4         &   \multicolumn{1}{c|}{ \textbf{\textcolor{darkred}{82.25\tiny{±0.13}}}}     &81.17\tiny{±1.18}&	81.56\tiny{±0.20}	&73.93\tiny{±4.04}	&79.40\tiny{±2.51}   & 79.68\tiny{±0.27}    & 79.12\tiny{±0.34} & 81.01\tiny{±0.62}   & \textbf{\textcolor{darkblue}{81.77\tiny{±0.32}}}  \\
 & \cellcolor{gray!12}Avg. Acc                                                                                      &                                                                        \multicolumn{1}{c|}{\cellcolor{gray!12} \textbf{\textcolor{darkred}{80.81\tiny{±1.94}}}}       &\cellcolor{gray!12}80.15\tiny{±2.23}&	\cellcolor{gray!12}79.81\tiny{±2.82}& \cellcolor{gray!12}73.93\tiny{±3.32}&	\cellcolor{gray!12}78.35\tiny{±2.14}                      & \cellcolor{gray!12}\textbf{\textcolor{darkblue}{80.76\tiny{±2.19}}}    & \cellcolor{gray!12}78.71\tiny{±2.67}                                                   & \cellcolor{gray!12}77.49\tiny{±2.45} & \cellcolor{gray!12}79.61\tiny{±2.15}                    \\
  & \cellcolor{gray!12}Avg. Rank                                                                                     &\multicolumn{1}{c|}{\cellcolor{gray!12}\textbf{\textcolor{darkred}{1.50}}}&\cellcolor{gray!12}5.00&\cellcolor{gray!12}6.50&		\cellcolor{gray!12}16.25&		\cellcolor{gray!12}8.75&	\cellcolor{gray!12}\textbf{\textcolor{darkblue}{2.75}}&		\cellcolor{gray!12}9.50&		\cellcolor{gray!12}12.50&		\cellcolor{gray!12}7.50	           \\ \midrule
\multicolumn{2}{c}{\cellcolor{green!8}Avg. All}                                                                                                                                            &     \cellcolor{green!8}75.47& \cellcolor{green!8}73.42 &\cellcolor{green!8}62.37 &\cellcolor{green!8}57.33   &\cellcolor{green!8}63.02 & \cellcolor{green!8}65.34 & \cellcolor{green!8}62.06 &\cellcolor{green!8}62.37& \cellcolor{green!8}71.43    \\
\multicolumn{2}{c} {\cellcolor{green!8}$p$-value}                                                                                                                                             &     \cellcolor{green!8}- & \cellcolor{green!8}\textbf{$\le$1e-4}  &  \cellcolor{green!8}\textbf{0.0001}  & \cellcolor{green!8}\textbf{$\le$1e-4}  &   \cellcolor{green!8}\textbf{$\le$1e-4} & \cellcolor{green!8}\textbf{$\le$1e-4} & \cellcolor{green!8} \textbf{$\le$1e-4}&  \cellcolor{green!8}\textbf{$\le$1e-4}  & \cellcolor{green!8}\textbf{0.0017}                                       \\  \midrule
\multicolumn{1}{c}{\multirow{2}{*}{Datasets}}                          & \multicolumn{1}{c}{\multirow{2}{*}{\begin{tabular}[c]{@{}c@{}}Target\\ Domain\end{tabular}}} & \multicolumn{5}{c|}{General OOD Generalization}                                                                                                                                                                                                                                                                                                                 & \multicolumn{4}{c}{General Time Series}                                                                                                                                                                                             \\  \cmidrule{3-11}
\multicolumn{1}{c}{}                                                   & \multicolumn{1}{c}{}                                                                         & \begin{tabular}[c]{@{}c@{}}GroupDRO\\ ICLR'20\end{tabular}             & \begin{tabular}[c]{@{}c@{}}VREx\\ ICML'21\end{tabular}        & \begin{tabular}[c]{@{}c@{}}SDL\\ AAAI '23\end{tabular} & \begin{tabular}[c]{@{}c@{}}FOOGD\\ NeurIPS'24\end{tabular}                 & \begin{tabular}[c]{@{}c@{}}EVIL\\ IJCV'25\end{tabular}                          & \begin{tabular}[c]{@{}c@{}}TARNet\\ KDD'21\end{tabular} & \begin{tabular}[c]{@{}c@{}}PatchTST\\ ICLR'23\end{tabular}   & \begin{tabular}[c]{@{}c@{}}TS-GAC\\ AAAI'25\end{tabular} & \begin{tabular}[c]{@{}c@{}}MPTSNet\\ AAAI'25\end{tabular}     \\ \midrule
\multirow{7}{*}{UCIHAR}                                                 & 0                                                                                             &  92.34\tiny{±1.24} &93.03\tiny{±0.88} & 85.25\tiny{±2.89} & 92.71\tiny{±2.05} &89.31\tiny{±2.88}&  73.43\tiny{±5.55} &94.31\tiny{±0.31} & 93.20\tiny{±0.81} &\textbf{\textcolor{darkblue}{94.51\tiny{±0.38}}}           \\
                                                                        & 1                                                                                             &   71.33\tiny{±1.80} & 72.92\tiny{±2.40} &61.60\tiny{±3.38} &80.07\tiny{±1.57} &70.82\tiny{±6.97} & 56.63\tiny{±2.80} &69.67\tiny{±2.08} &73.32\tiny{±3.68} &78.08\tiny{±0.93}   \\
                                                                        & 2                                                                                             &   92.32\tiny{±0.77} &92.38\tiny{±1.08} &92.49\tiny{±1.05} &93.30\tiny{±2.00}   &  85.73\tiny{±0.73} &92.03\tiny{±3.28} &94.30\tiny{±0.39} &93.61\tiny{±1.49} & 93.03\tiny{±1.27}  \\
                                                                        & 3                                                                                             & 86.69\tiny{±5.15} &87.01\tiny{±4.15} &75.34\tiny{±8.72} &85.98\tiny{±4.70} &83.95\tiny{±3.76} &93.44\tiny{±4.76} &87.76\tiny{±1.51} &90.29\tiny{±2.94} &87.20\tiny{±0.66}    \\
                                                                        & 4                                                                                             &  94.24\tiny{±5.49} &93.65\tiny{±4.45} &82.59\tiny{±1.28} &91.70\tiny{±2.40} &92.16\tiny{±3.42} &77.68\tiny{±2.83} &93.58\tiny{±3.57} & 94.77\tiny{±0.60} &\textbf{\textcolor{darkblue}{96.30\tiny{±1.23}}}   \\
                                                                        & \cellcolor{gray!12}Avg. Acc                                                                                      &   \cellcolor{gray!12}87.32\tiny{±9.14} &\cellcolor{gray!12}87.79\tiny{±8.41} &\cellcolor{gray!12}79.43\tiny{±11.42} &\cellcolor{gray!12}88.59\tiny{±5.59} & \cellcolor{gray!12}84.39\tiny{±7.82}      &                                                                \cellcolor{gray!12}78.64\tiny{±14.26} & \cellcolor{gray!12}87.92\tiny{±9.82} & \cellcolor{gray!12}89.04\tiny{±8.42} & \cellcolor{gray!12}89.82\tiny{±6.81}    \\
                                                                        & \cellcolor{gray!12}Avg. Rank                                                                                     &                                                                        \cellcolor{gray!12}9.40&        \cellcolor{gray!12}8.40                                                     &   \cellcolor{gray!12}13.60       & \cellcolor{gray!12}8.00                                                                          &             \cellcolor{gray!12}10.20   & \cellcolor{gray!12}13.20                                                        &     \cellcolor{gray!12}6.00&              \cellcolor{gray!12}5.60    &           \cellcolor{gray!12}4.80                                                \\ \midrule
\multirow{6}{*}{\begin{tabular}[c]{@{}c@{}}UniMiB-\\ SHAR\end{tabular}} & 1                                                                                             &      43.91\tiny{±1.86} &43.91\tiny{±2.27} &42.71\tiny{±2.41} & 55.02\tiny{±7.13} & 45.37\tiny{±1.50}&         36.15\tiny{±3.81} & 45.26\tiny{±1.56} &44.74\tiny{±1.07} & 46.36\tiny{±1.05}         \\
                                                                        & 2                                                                                             & 38.60\tiny{±1.60} &39.87\tiny{±1.38} & 41.48\tiny{±2.36} &47.99\tiny{±3.11} &45.42\tiny{±3.01}& 34.17\tiny{±7.60} &42.89\tiny{±1.63} &42.99\tiny{±0.78} & 47.14\tiny{±2.29}    \\
                                                                        & 3                                                                                             &                                                                         64.28\tiny{±1.66} &65.92\tiny{±1.18} &64.68\tiny{±0.92} &66.57\tiny{±4.30} &64.53\tiny{±2.81} & 42.77\tiny{±3.91} & 69.08\tiny{±1.32} &65.99\tiny{±3.02} & 70.07\tiny{±1.46}   \\
                                                                        & 5                                                                                             &                                                                         39.67\tiny{±0.61} & 39.87\tiny{±1.57} &38.73\tiny{±1.88} & \textbf{\textcolor{darkred}{43.41\tiny{±5.83}}} & 39.28\tiny{±2.88} &33.50\tiny{±2.21} &39.94\tiny{±0.63} &39.47\tiny{±0.66} &41.28\tiny{±1.09}   \\
                                                                        & \cellcolor{gray!12}Avg. Acc                                                                                      &  \cellcolor{gray!12}46.61\tiny{±10.75} &\cellcolor{gray!12}47.39\tiny{±11.21} &\cellcolor{gray!12}46.90\tiny{±10.79} &\cellcolor{gray!12}53.25\tiny{±10.21} &\cellcolor{gray!12}48.65\tiny{±10.04}   &\cellcolor{gray!12}36.65\tiny{±5.80} &\cellcolor{gray!12}49.29\tiny{±11.95} &\cellcolor{gray!12}48.30\tiny{±10.78} & \cellcolor{gray!12}51.21\tiny{±11.50}     \\
                                                                        & \cellcolor{gray!12}Avg. Rank                                                                                     &   \cellcolor{gray!12}11.00	& \cellcolor{gray!12}10.00	& \cellcolor{gray!12}11.25& \cellcolor{gray!12}\textbf{\textcolor{darkblue}{3.25}}	& \cellcolor{gray!12}9.00	& \cellcolor{gray!12}15.25	& \cellcolor{gray!12}7.50	& \cellcolor{gray!12}9.00		& \cellcolor{gray!12}5.25       \\ \midrule
\multirow{6}{*}{EMG}                                                    & 0                                                                                             &                                                                        44.50\tiny{±3.24} &45.03\tiny{±3.11} & 43.96\tiny{±3.52} & 56.93\tiny{±8.52} &61.63\tiny{±8.25}  &50.59\tiny{±2.92} & 48.24\tiny{±1.04} &48.88\tiny{±0.48} & 43.86\tiny{±0.54}    \\
                                                                        & 1                                                                                             &                                                                         38.54\tiny{±1.95} &38.64\tiny{±1.87} & 39.59\tiny{±1.98} &45.88\tiny{±7.17} & 47.38\tiny{±5.30}  &48.46\tiny{±1.47} & 43.78\tiny{±1.03} &43.04\tiny{±1.76} & 41.57\tiny{±2.56}   \\
                                                                        & 2                                                                                             &   39.31\tiny{±1.52} &39.30\tiny{±0.67} &38.26\tiny{±0.76} &59.84\tiny{±7.72} &51.68\tiny{±6.21} &44.77\tiny{±2.06} &40.00\tiny{±1.62} &41.51\tiny{±1.46} &51.52\tiny{±1.58}    \\
                                                                        & 3                                                                                             &                                                                        49.43\tiny{±2.56} &48.37\tiny{±2.30} & 47.79\tiny{±1.39} & 77.91\tiny{±5.78} &60.40\tiny{±8.44} & 57.50\tiny{±1.11} &51.45\tiny{±1.60} &52.51\tiny{±1.43} &51.64\tiny{±0.88}    \\
                                                                        & \cellcolor{gray!12}Avg. Acc                                                                                      &    \cellcolor{gray!12}42.94\tiny{±5.02} & \cellcolor{gray!12}42.84\tiny{±4.61} &\cellcolor{gray!12}42.40\tiny{±4.35} &\cellcolor{gray!12}60.14\tiny{±13.61} & \cellcolor{gray!12}55.27\tiny{±8.99}  &\cellcolor{gray!12}50.33\tiny{±5.10} &\cellcolor{gray!12}45.87\tiny{±4.63} &\cellcolor{gray!12}46.48\tiny{±4.72} & \cellcolor{gray!12}47.15\tiny{±4.85}    \\
                                                                        &\cellcolor{gray!12}Avg. Rank                                                                                     &\cellcolor{gray!12}15.25	&\cellcolor{gray!12}15.00	&\cellcolor{gray!12}15.50	&	\cellcolor{gray!12}5.25&	\cellcolor{gray!12}5.75&\cellcolor{gray!12}8.00	&\cellcolor{gray!12}12.00		&\cellcolor{gray!12}11.00		&\cellcolor{gray!12}12.00          \\ \midrule
\multirow{6}{*}{\begin{tabular}[c]{@{}c@{}}Opportu\\nity\end{tabular}} & S1                                                                                            &                                                   74.84\tiny{±2.67} &75.30\tiny{±2.46} &75.01\tiny{±0.57} &81.31\tiny{±1.31} & 72.90\tiny{±7.66}                       &72.37\tiny{±0.01} & \textbf{\textcolor{darkred}{83.39\tiny{±0.25}}} &81.79\tiny{±0.20} &80.35\tiny{±0.29}                                \\
                                                                        & S2                                                                                            &                                                                        74.95\tiny{±1.86} & 74.90\tiny{±1.78} & 73.51\tiny{±0.85} &79.91\tiny{±1.22} &74.03\tiny{±4.94} &71.98\tiny{±0.01} &81.01\tiny{±0.62} & 80.92\tiny{±0.15} & 79.25\tiny{±0.44}       \\
                                                                        & S3                                                                                            &                                                                         75.98\tiny{±0.55} & 73.80\tiny{±1.46} & 75.11\tiny{±0.27} & 75.18\tiny{±3.31} &74.59\tiny{±2.89} &71.49\tiny{±0.01} &75.98\tiny{±0.93} & 75.40\tiny{±0.89} & 76.53\tiny{±1.14}     \\
                                                                        & S4                                                                                            &                                                                        75.97\tiny{±0.60} &74.97\tiny{±0.30} & 78.42\tiny{±3.73} & 77.75\tiny{±3.47}   &81.32\tiny{±0.65} & 73.40\tiny{±0.00} & 81.37\tiny{±0.55} & \textbf{\textcolor{darkblue}{81.77\tiny{±0.32}}} &79.27\tiny{±1.01}   \\
                                                                        & \cellcolor{gray!12}Avg. Acc                                                                                      &\cellcolor{gray!12}75.39\tiny{±1.65} &\cellcolor{gray!12}74.99\tiny{±1.77} & \cellcolor{gray!12}74.65\tiny{±0.85} &\cellcolor{gray!12}78.70\tiny{±3.37} &\cellcolor{gray!12}74.82\tiny{±5.02}&\cellcolor{gray!12}72.31\tiny{±0.73} & \cellcolor{gray!12}80.59\tiny{±2.90} & \cellcolor{gray!12}79.97\tiny{±2.77} & \cellcolor{gray!12}78.85\tiny{±1.63}    \\
                                                                        & \cellcolor{gray!12}Avg. Rank                                                                                     &\cellcolor{gray!12}12.50&\cellcolor{gray!12}14.00	&	\cellcolor{gray!12}14.50		&\cellcolor{gray!12}9.25		&\cellcolor{gray!12} 14.75	&\cellcolor{gray!12}18.00	&	\cellcolor{gray!12}4.00	&	\cellcolor{gray!12}5.00	&	\cellcolor{gray!12}8.75    \\ \midrule
\multicolumn{2}{c}{ \cellcolor{green!8}Avg. All}                                                                                                                                            &   \cellcolor{green!8}64.49 & \cellcolor{green!8}64.69& \cellcolor{green!8}61.93 & \cellcolor{green!8}71.25&\cellcolor{green!8}67.32 &\cellcolor{green!8}60.61 & \cellcolor{green!8}67.17 & \cellcolor{green!8}67.30 & \cellcolor{green!8}68.11       \\ 
\multicolumn{2}{c}{\cellcolor{green!8}$p$-value}                                                                                                                                             &     \cellcolor{green!8}\textbf{$\le$1e-4} &\cellcolor{green!8}\textbf{$\le$1e-4}  & \cellcolor{green!8}\textbf{$\le$1e-4}  & \cellcolor{green!8}\textbf{$\le$1e-4}   & \cellcolor{green!8}\textbf{$\le$1e-4} &\cellcolor{green!8}\textbf{$\le$1e-4} &\cellcolor{green!8}\textbf{$\le$1e-4}   & \cellcolor{green!8}\textbf{$\le$1e-4}  & \cellcolor{green!8}\textbf{$\le$1e-4}                                                  \\ \bottomrule
\end{tabular}
\label{tab_main}
\end{table*}

\begin{table}
\centering
\setlength{\tabcolsep}{0.8mm}
\caption{Performance comparison of energy-based methods. The result is the average of all target domains on each dataset. ECE stands for expected calibration error ($\downarrow$). The $p$-value is obtained using the Wilcoxon signed-rank test.}
\begin{tabular}{c|c|c|c|c|c}
\toprule
\multicolumn{1}{c}{Datasets} & \multicolumn{1}{c}{\begin{tabular}[c]{@{}c@{}}ERIS\\ (Ours)\end{tabular}} & \multicolumn{1}{c}{\begin{tabular}[c]{@{}c@{}}EOW-Softmax\\ ICCV'21\end{tabular}} & \multicolumn{1}{c}{\begin{tabular}[c]{@{}c@{}}ELI\\ CVPR'22\end{tabular}} & \multicolumn{1}{c}{\begin{tabular}[c]{@{}c@{}}EGC\\ ICCV‘23\end{tabular}} & \multicolumn{1}{c}{\begin{tabular}[c]{@{}c@{}}Golan\\ NeurIPS'24\end{tabular}}\\ \midrule
\multirow{1}{*}{UCIHAR}        & \textbf{0.07\tiny{±0.03}} 
                            &0.09\tiny{±0.04}         &0.43\tiny{±0.06}        &0.45\tiny{±0.09}                & 0.26\tiny{±0.11}     \\
\multirow{1}{*}{\shortstack{UniMiB-SHAR}}  
& \textbf{0.17\tiny{±0.11}}  & 0.25\tiny{±0.12}      &0.21\tiny{±0.05}      & 0.37\tiny{±0.13}    &0.21\tiny{±0.12}        \\ 
\multirow{1}{*}{EMG}          
 & \textbf{0.20\tiny{±0.12}}           &0.26\tiny{±0.13}       & 0.21\tiny{±0.07}  &0.41\tiny{±0.14}    &0.28\tiny{±0.08}    \\  
\multirow{1}{*}{Opportunity}  
 & \textbf{0.06\tiny{±0.03}}                   & 0.08\tiny{±0.03}             & 0.12\tiny{±0.02}            & 0.29\tiny{±0.16}                    &0.12\tiny{±0.03}              \\  \midrule
 \multicolumn{1}{c|}{\cellcolor{green!8}Avg. ALL}  &\cellcolor{green!8}\textbf{0.12} &\cellcolor{green!8}0.17 &\cellcolor{green!8}0.26 &\cellcolor{green!8}0.39&\cellcolor{green!8}0.22 \\
 \multicolumn{1}{c|}{\cellcolor{green!8}$p$-value} & \cellcolor{green!8}-&     \cellcolor{green!8}\textbf{0.0001}  &  \cellcolor{green!8}\textbf{$\le$1e-4}   & \cellcolor{green!8} \textbf{$\le$1e-4}  & \cellcolor{green!8} \textbf{$\le$1e-4}  \\
 \bottomrule
\end{tabular}
\label{tab_energy}
\end{table}

\subsubsection{Evaluation Metrics} We comprehensively evaluate the OOD generalization performance of the proposed model and all baseline models on four public benchmark datasets. To ensure robustness, each experiment is repeated independently five times, and we primarily report the mean accuracy (ACC, in \%) and standard deviation. In comparisons with energy-based methods, we also include the Expected Calibration Error (ECE) to evaluate the reliability of model confidence estimates. A lower ECE indicates better calibration and helps mitigate overconfidence. Furthermore, ablation studies provide additional results using the F1-score, precision, and recall (\%). All comparisons report average rankings, and statistical significance is analyzed using the Wilcoxon signed-rank test.

\subsubsection{Implementation Details} All experiments conducte on a server equipped with an Intel Xeon Gold 6346 CPU and two NVIDIA GeForce RTX 3090 GPUs, each with 24GB of memory. The server runs Ubuntu 22.04.1, and the software environment is built on CUDA 12.8 and PyTorch 2.7.0. We primarily adopte the hyperparameter configurations from the baseline methods, with necessary adjustments made to optimize performance based on the characteristics of each dataset. Models are trained for 100 epochs with a fixed batch size of 64. We employe the Adam optimizer with an initial learning rate of 1e-4 and a weight decay of 1e-5. A StepLR scheduler is used to decay the learning rate by a factor of 0.1 every 30 epochs. Notably, the adversarial strength is dynamically adjusted throughout the training process. To ensure the robustness and reproducibility of our results, all experiments are conducted with a fixed random seed and run in parallel on both GPUs. Each experiment is repeated five times. Full implementation and details are in the repository\footnote{\textcolor{magenta}{\url{https://wuliwuxin.github.io/ERISProject}}}.

\subsection{Main Results}
\label{ex_main}
\textbf{Overall Performance.} The superior capabilities of the ERIS model are quantitatively substantiated through a comprehensive analysis of its generalization performance and model calibration, with detailed results presented in  \ref{tab_main} and Table \ref{tab_energy}, respectively. \underline{\textbf{Firstly}}, in terms of generalization, Table \ref{tab_main} demonstrates that ERIS significantly outperforms a comprehensive suite of state-of-the-art (SOTA) methods. This extensive comparison spans four distinct categories of baselines: energy-based methods, general time series methods, general OOD methods, and time series OOD methods. Across this broad evaluation, ERIS achieves the highest overall average accuracy of 75.47\% and secures the best average rank of 1.375. \underline{\textbf{Secondly}}, the model demonstrates exceptional adaptability. Its consistently high performance across four distinct and challenging real-world datasets as detailed in Table \ref{tab_main}, highlights its capacity to generalize effectively across heterogeneous data domains. \underline{\textbf{Thirdly}}, beyond raw predictive accuracy, ERIS excels in model calibration. When compared against other energy-based models in Table \ref{tab_energy}, it achieves the lowest average ECE of 0.12. This result is critical, as a lower ECE signifies a more reliable alignment between the model's predictive confidence and its actual likelihood of being correct, thereby mitigating the risk of overconfident predictions in OOD scenarios. \underline{\textbf{Fourthly}}, the superiority of ERIS is statistically robust. The Wilcoxon signed-rank test confirmed that the performance gains over all compared methods are statistically significant (with \textbf{all $p$-values$<$0.05}), affirming that these results are not attributable to random chance. \underline{\textbf{Finally}}, ERIS proves to be both effective and stable. The minimal fluctuation in its performance, evidenced by its consistently high rankings in Table \ref{tab_main} and Table \ref{tab_energy}, underscores the reliability and robustness of our proposed framework.

\begin{figure}
\centering
\includegraphics[width=1\columnwidth]{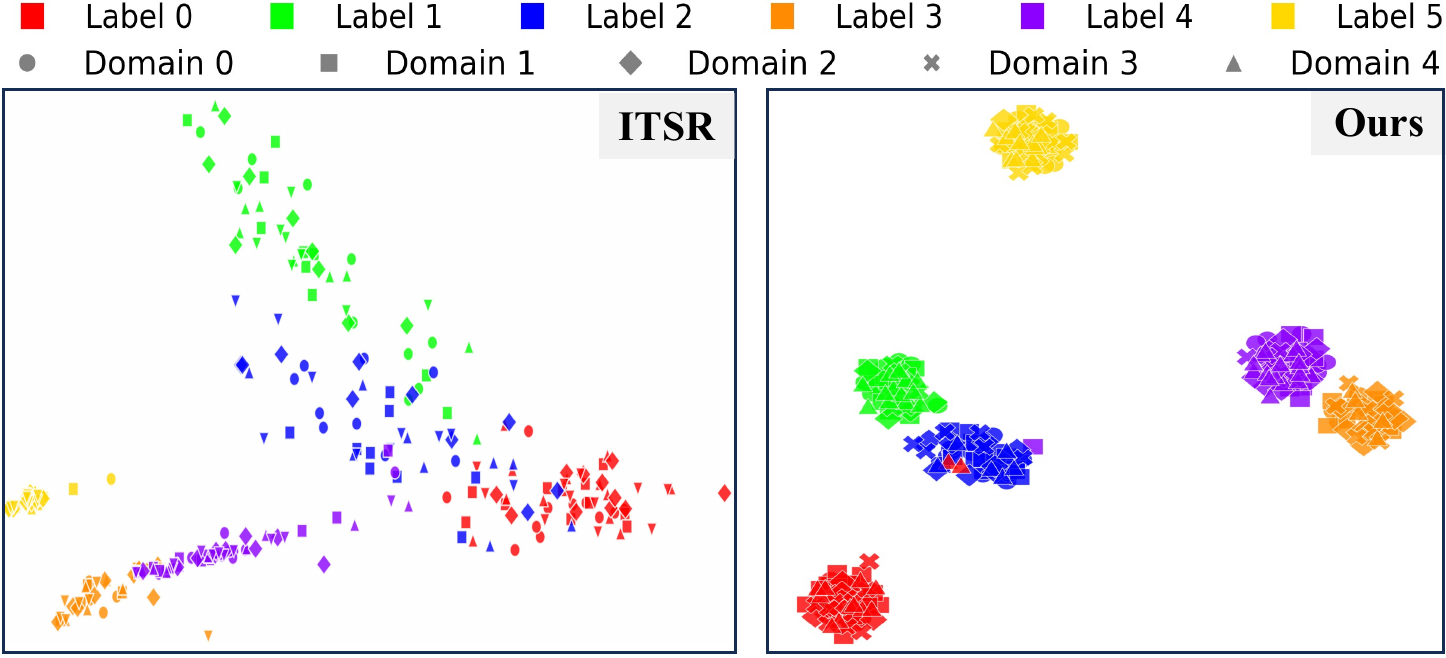}
\caption{Comparison of t-SNE visualization results of ITSR and ours ERIS on the UCIHAR dataset. Each color denotes a different activity label, while different marker shapes correspond to different domains.}
\label{fig_tsen}
\end{figure}

\textbf{Visualization.} To qualitatively assess the feature disentanglement and invariance capabilities of our model, we use t-SNE to visualize the feature representations learned by ERIS (guided method) and ITSR (unguided method), on the UCIHAR dataset. The results, presented in Fig. \ref{fig_tsen}, reveal a clear distinction in representation quality. The features from ITSR (left plot) show a high degree of entanglement; while data points are loosely grouped by label (color), the clusters are poorly defined, elongated, and exhibit significant overlap. This indicates that domain-specific variations are still entangled with label-relevant information, resulting in ambiguous class boundaries. In stark contrast, ERIS (right plot) produces highly compact and well-separated clusters for each label, demonstrating superior discriminative power. More importantly, within each of these tight label-specific clusters, points from all different domains are collapsed together. This visually confirms that ERIS has successfully learned a domain-invariant representation by effectively discarding domain-specific features while preserving only the essential class information. This strong qualitative evidence indicates that ERIS has a superior capability in handling the heterogeneity between different domains, leading to a more robust and generalizable feature space.

\subsection{Ablation Studies}
\label{sec:ab}

\begin{figure*}
\centering
\includegraphics[width=1\textwidth]{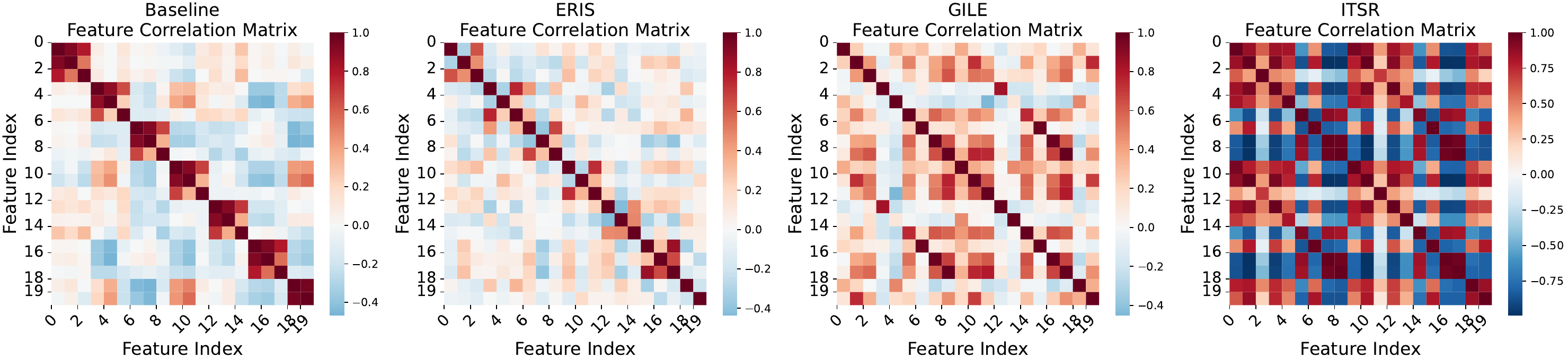} 
\caption{Feature correlation matrices for different orthogonal methods on the UCIHAR dataset visualize pairwise feature correlations. Red shows positive correlations (darker for stronger), blue indicates negative correlations, and white represents no correlation.}
\label{fig_orl}
\end{figure*}

\begin{figure*}
\includegraphics[width=1\textwidth]{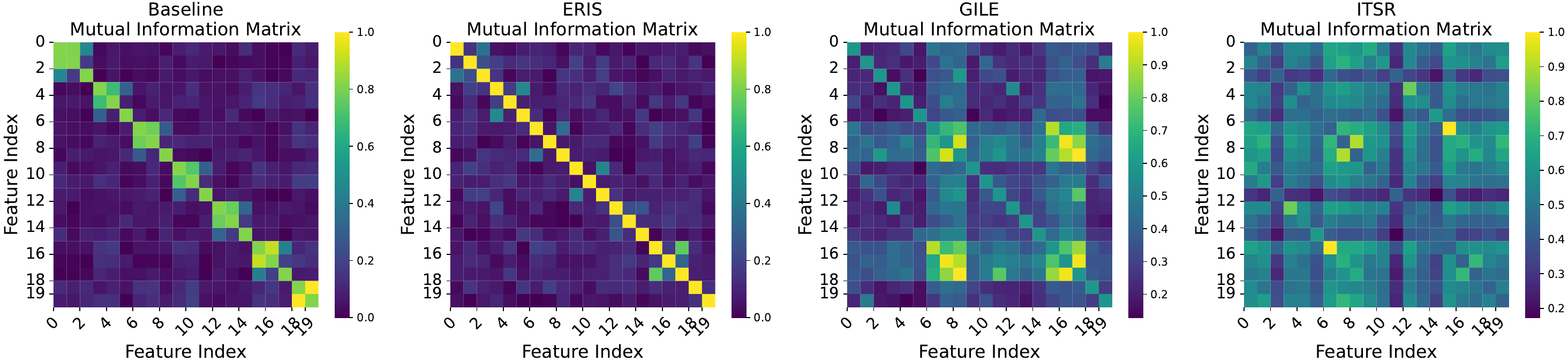} 
\caption{Visualizing orthogonal mutual information matrices. If two features are independent, their mutual information is zero. The higher the mutual information, the more information these two features share, meaning that knowing one feature reduces the uncertainty about the other.}
\label{fig_mi}
\end{figure*}

\begin{figure}
\centering
\includegraphics[width=0.45\textwidth]{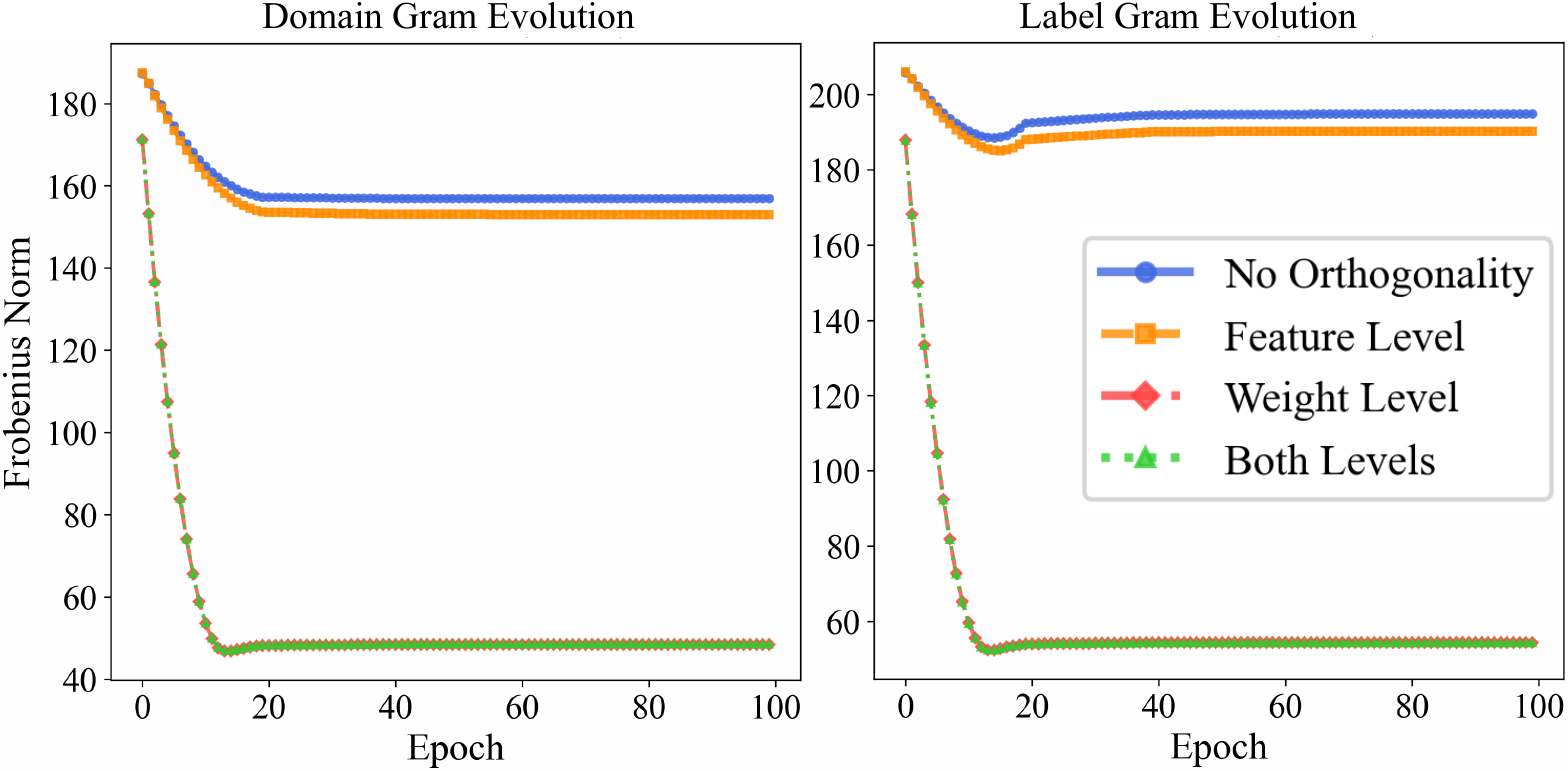} 
\caption{Impact of various orthogonality strategies on the Frobenius norm of model parameters during ERIS training on the UCIHAR dataset.}
\label{fig_norms}
\end{figure}

\textbf{How do the orthogonality methods affect the performance?} To provide a multi-faceted evaluation of how different orthogonality methods affect performance, we analyze their impact on both the final feature representations and the underlying model parameters. \underline{\textbf{Firstly}}, a qualitative analysis of the feature correlation matrices in Fig. \ref{fig_orl} offers visual proof of our method's superiority. Compared to the TSMixer baseline, ERIS produces a significantly clearer and more dispersed matrix with correlation coefficients predominantly concentrated near zero, indicating effective removal of linear dependencies. In contrast, alternative methods show clear deficiencies: GILE \cite{qian2021} presents an overly concentrated block structure, while ITSR \cite{shi2024} exhibits a rigid "checkerboard" pattern, suggesting it may over-constrain feature relationships. \underline{\textbf{Secondly}}, building on this, the mutual information (MI) analysis in Fig. \ref{fig_mi}, which quantifies both linear and non-linear dependencies, offers deeper insight. The baseline model exhibits pronounced off-diagonal intensities, indicating substantial data redundancy. While GILE and ITSR partially attenuate these dependencies, they still leave behind scattered regions of high MI, showing that spurious non-linear correlations are incompletely removed. By contrast, ERIS yields MI values that are markedly reduced off-diagonally, demonstrating that redundant inter-feature information has been substantially eliminated to create a more compact and independent representation. \underline{\textbf{Thirdly}}, this qualitative superiority is substantiated by a quantitative analysis of the model's parameters in Fig. \ref{fig_norms}. Without any constraint, the Frobenius norm of the projection matrices stabilizes at a high level, indicative of entangled parameter subspaces. Notably, applying a conventional feature-level constraint fails to alter this outcome, with its norm evolution curve remaining nearly identical to the baseline. This confirms that such local, "per-instance" constraints are insufficient to enforce global structural separation. In stark contrast, ERIS’s weight-level strategy fundamentally alters this dynamic. By directly minimizing the $\mathcal{L}_\mathrm{ortho}$ penalty on the parameter matrices ($\boldsymbol{W_d}$ and $\boldsymbol{W_l}$), our approach compels the Frobenius norm to decrease sharply and converge to a minimal value. This global, structural constraint effectively enforces independence between the domain and label subspaces, which is the mechanism behind the feature disentanglement observed.

\begin{figure}
\centering
\includegraphics[width=0.95\columnwidth]{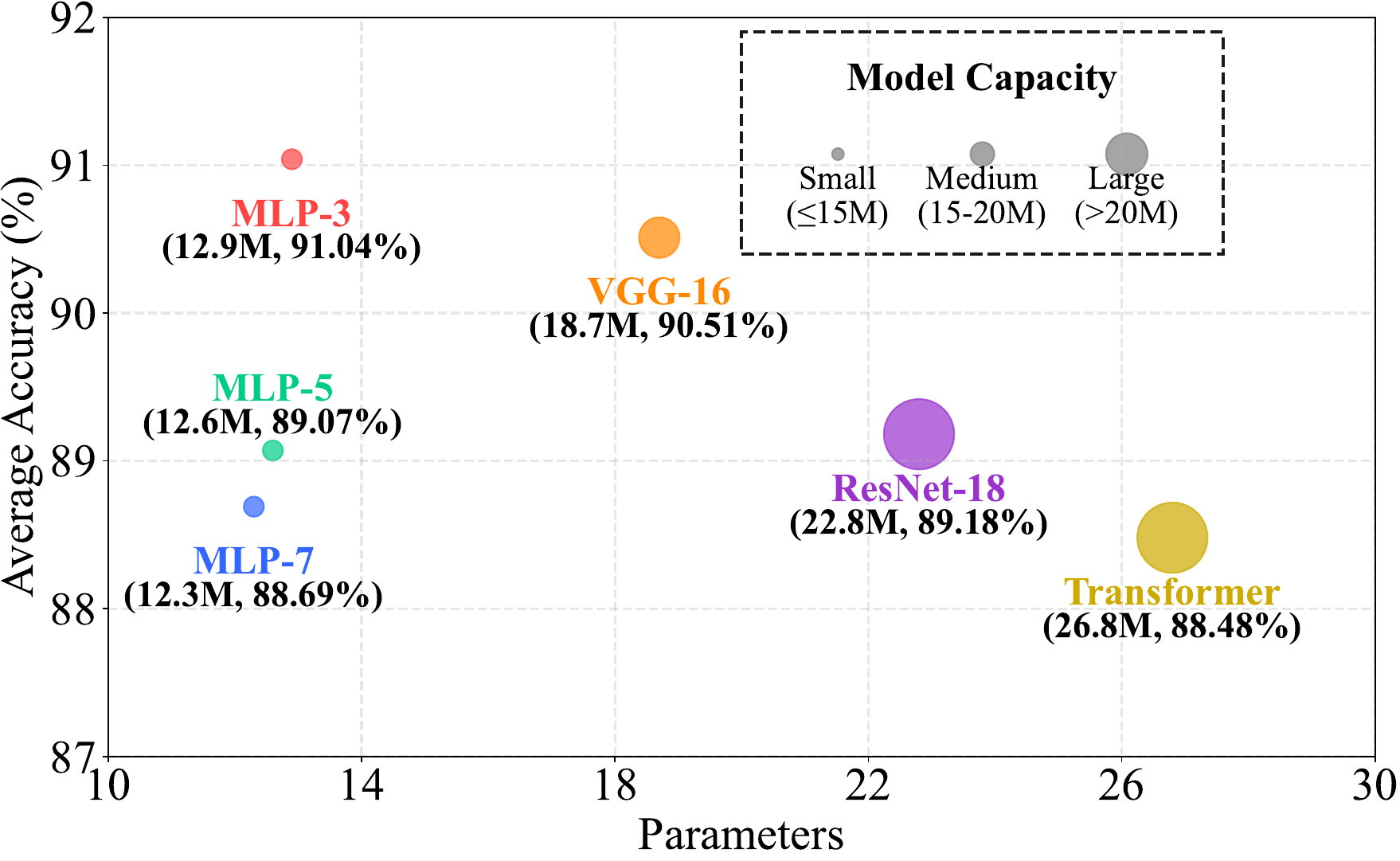} 
\caption{The impact of model capacity and architecture on the classification accuracy of the ERIS framework on the UCIHAR dataset. The size of the bubbles corresponds to the parameter scale of the models.}
\label{fig_bubble}
\end{figure}

\textbf{How does the energy-guided mechanism capacity affect the performance?} The core finding of this study is that the performance of the ERIS framework is not merely dictated by model capacity, but rather stems from the effective synergy between its unique energy-guided mechanism and various model architectures. \underline{\textbf{Firstly}}, model capacity does not correlate directly with performance. As illustrated in Fig. \ref{fig_bubble}, a small-capacity MLP-3 model (12.9M) achieves the highest accuracy of 91.04\%, outperforming significantly larger models like ResNet-18 (22.8M) and Transformer (26.8M). This result indicates that the energy guidance and orthogonality constraints within ERIS provide effective semantic direction and structural regularization. \underline{\textbf{Secondly}}, the ERIS framework demonstrates cross-architecture robustness. Table \ref{tab_detailed_analysis} reveals that ERIS consistently delivers stable performance gains across MLP, CNN, and Transformer backbones. Notably, with the Transformer architecture where other methods cause significant performance degradation, ERIS maintains a high level of accuracy. In conclusion, the key to ERIS's effectiveness lies in its energy-guided mechanism, not model capacity. By providing semantic guidance and structural constraints, this mechanism empowers diverse models to efficiently learn invariant features, thereby achieving superior generalization performance.

\begin{table}[htbp]
\setlength{\tabcolsep}{0.8mm}
\centering
\caption{Comparative Performance Evaluation of Different Training Paradigms. Architectures: MLP (TSMixer \cite{chen2023tsmixer}), CNN (TCN \cite{bai2018tcn}), Transformer (TimeXer\cite{Wang2024TimeXer}). Values represent classification accuracy (\%); $\Delta$(\%) indicates relative change from baseline.}
\label{tab_detailed_analysis}
\footnotesize
\begin{tabular}{c|c|c|cc|cc|cc}
\toprule
\multirow{2}{*}{Dataset} & \multirow{2}{*}{Model} & \multirow{2}{*}{Baseline} & 
\multicolumn{2}{c|}{GILE} & \multicolumn{2}{c|}{ITSR} & \multicolumn{2}{c}{Ours} \\
\cmidrule(lr){4-5} \cmidrule(lr){6-7} \cmidrule(lr){8-9}
& & & Perf. & $\Delta$(\%) & Perf. & $\Delta$(\%) & Perf. & $\Delta$(\%) \\
\midrule
\multirow{3}{*}{UCIHAR}
& MLP& 87.6 & 75.5 & \textcolor{red}{-13.8} & 83.9 & \textcolor{green!70!black}{+0.2} & 89.5 & \textcolor{green!70!black}{+2.1} \\
& CNN & 83.7 & 83.0 & \textcolor{red}{-4.9} & 83.1 & \textcolor{red}{-4.8} & 87.5 & \textcolor{green!70!black}{+4.5} \\
& Transformer& 87.2 & 83.4 & \textcolor{red}{-4.9} & 77.2 & \textcolor{red}{-7.8} & 85.7 & \textcolor{red}{-1.7} \\
\midrule
\multirow{3}{*}{\begin{tabular}[c]{@{}c@{}}UniMiB-\\ SHAR\end{tabular}} 
& MLP & 54.1 & 47.6 & \textcolor{red}{-12.0} & 53.0 & \textcolor{red}{-2.0} & 55.9 & \textcolor{green!70!black}{+3.3} \\
& CNN& 52.3 & 43.4 & \textcolor{red}{-17.0} & 39.5 & \textcolor{red}{-24.5} & 51.3 & \textcolor{red}{-1.9} \\
& Transformer & 51.6 & 45.7 & \textcolor{red}{-11.4} & 44.6 & \textcolor{red}{-13.6} & 47.7 & \textcolor{red}{-7.6} \\
\bottomrule
\end{tabular}
\end{table}

\begin{figure*}
\centering
\includegraphics[width=1.95\columnwidth]{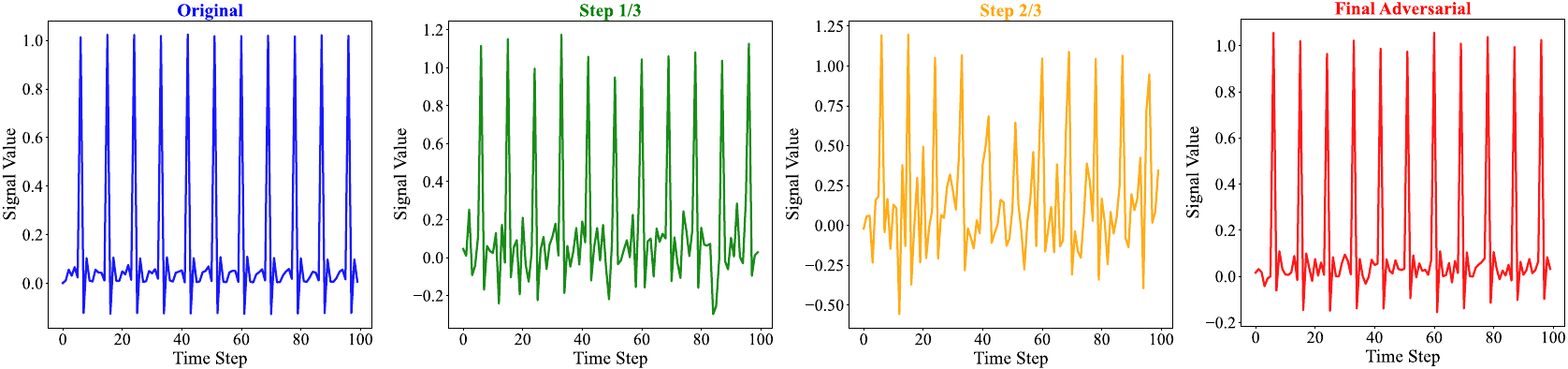} 
\caption{Visualization of the iterative adversarial sample generation on the UCIHAR dataset. The process begins with an "Original" signal , where the AG mechanism incrementally applies optimized, structured perturbations ("Step 1/3" , "Step 2/3" ) to produce a "Final Adversarial" sample.}
\label{fig_example}
\end{figure*}

\textbf{How does the AG mechanism enhance local robustness?} To visually validate the effectiveness of the AG mechanism in enhancing local feature robustness, Fig. \ref{fig_example} illustrates the iterative perturbation process applied to a sample signal from the UCIHAR dataset. The process commences with an "Original" signal. Subsequently, guided by Eq. \ref{eq:adv}, the AG mechanism computes a perturbation via iterative optimization to maximize the KL-divergence between the model's predictive distributions on the perturbed and original samples. The intermediate steps ("Step 1/3" and "Step 2/3" ) reveal how this gradient-based, structured perturbation is progressively injected, thereby altering the signal's morphology. Unlike random noise, this perturbation is specifically crafted within a given norm to be maximally confusing to the model. The "Final Adversarial" sample is the culmination of this optimization, representing a challenging instance situated near the model's decision boundary. By minimizing $\mathcal{L}_\mathrm{reg}$, the model is trained to maintain predictive consistency for both original and adversarial samples, a process that effectively smooths the latent feature space. 

\begin{figure}
\centering
\includegraphics[width=0.95\columnwidth]{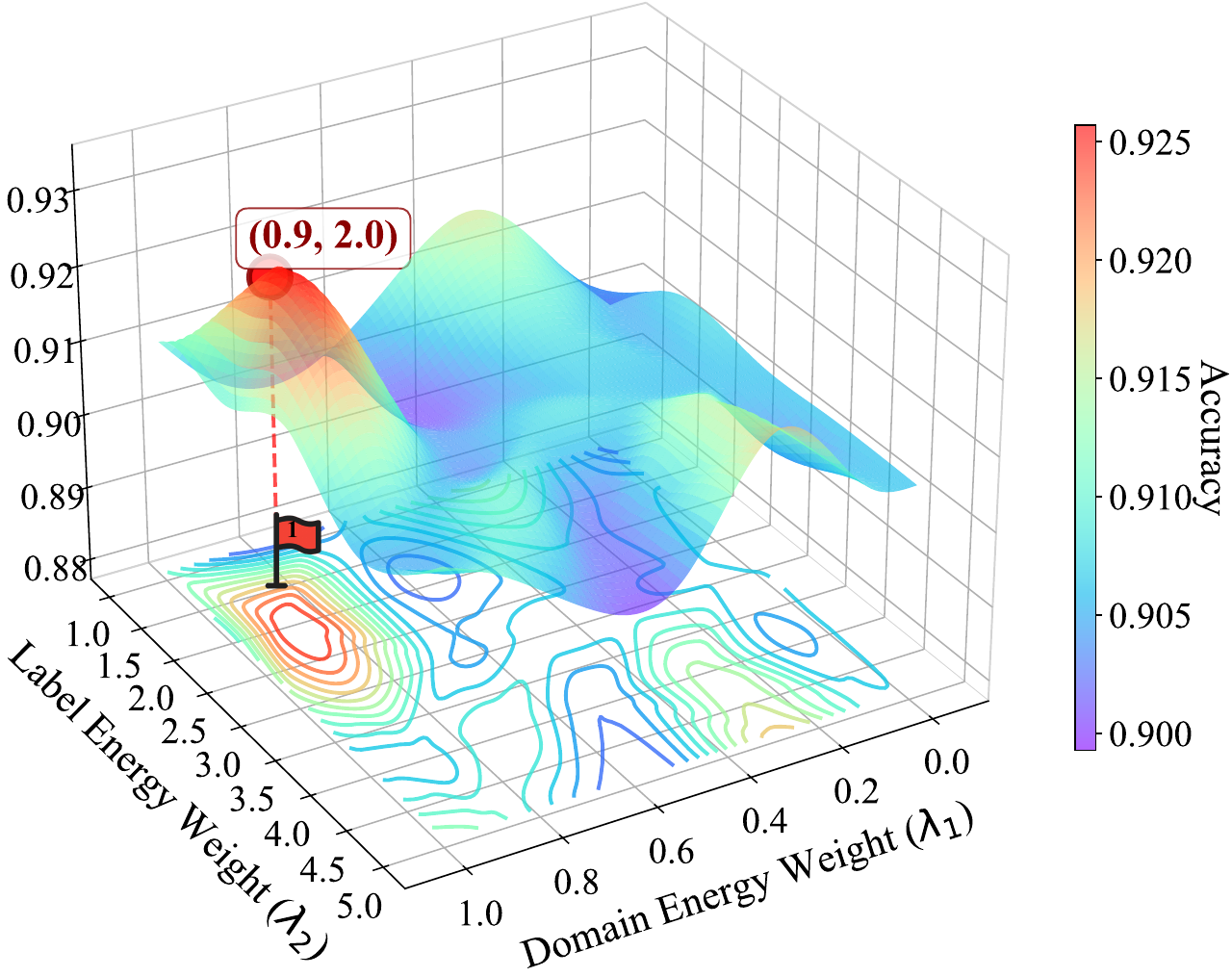} 
\caption{Ablation study of  the parameters $\lambda _{1}$ and $\lambda_{2}$ on the UCIHAR dataset.}
\label{fig_ab}
\end{figure}

\textbf{How does the loss weighting factor affect the performance?} The performance of the ERIS model is sensitive to the trade-off between the $\mathcal{L}_\mathrm{DSE}$ and the $\mathcal{L}_\mathrm{LSE}$, which are balanced by the hyperparameters $\lambda_{1}$ and $\lambda_{2}$ in the total loss function (Eq. \ref{eq:total}). We conducted experiments on the UCIHAR dataset to analyze this relationship, with the results visualized in the 3D surface plot in Fig. \ref{fig_ab}. The plot empirically demonstrates that the optimal performance is achieved with a configuration where $\lambda_{2}$ is significantly larger than $\lambda_{1}$. Specifically, the peak accuracy is reached when $\lambda_{1}$ is set to 0.9 and $\lambda_{2}$ is set to 2.0. The rationale behind this configuration is that a larger $\lambda_{2}$  strongly guides the model to prioritize the LSE process, which is driven by class prototypes. This encourages the learned features to form compact clusters around their correct prototypes while simultaneously pushing them away from others, thereby creating clear and discriminative class boundaries. While the label-centric task is prioritized, a substantial weight for $\lambda_{1}$ (0.9) remains crucial, as it ensures the model actively disentangles domain-specific variations. 

\begin{table}
\centering
\caption{Ablation results of ours ERIS. All metrics are averaged across five target domains of the UCIHAR dataset.}
\setlength{\tabcolsep}{0.8mm}
\begin{tabular}{c|ccc|cccc}
\toprule 
Index          & \textbf{DSE} & \textbf{LSE} & \textbf{AR} & ACC   &F1   & Precision & Recall \\    \midrule 
A          & \Checkmark  & \textcolor[cmyk]{0,0,0,.247}  {\XSolidBrush}   & \textcolor[cmyk]{0,0,0,.247}  {\XSolidBrush}   & 80.50 & 79.30 & 84.18     & 80.30  \\
B       &\textcolor[cmyk]{0,0,0,.247}  {\XSolidBrush} &\Checkmark            & \textcolor[cmyk]{0,0,0,.247}  {\XSolidBrush}  &81.81 & 80.80 & 84.64     &81.58  \\
C  (w/o $\mathcal{L}_{\mathrm{ortho}}$)        & \Checkmark & \Checkmark& \textcolor[cmyk]{0,0,0,.247}  {\XSolidBrush}   & 84.16 & 82.88 &87.43     & 83.71  \\
D   (w/ $\mathcal{L}_{\mathrm{ortho}}$)       &\Checkmark &\Checkmark & \textcolor[cmyk]{0,0,0,.247}  {\XSolidBrush} & 85.69 &86.50 &88.80 &87.95 \\
E& \textcolor[cmyk]{0,0,0,.247}  {\XSolidBrush}            & \Checkmark  & \Checkmark& 85.33 &84.17 &88.22    & 85.13  \\  \midrule 
F (w/o $\mathcal{L}_{\mathrm{ortho}}$) & \Checkmark & \Checkmark  & \Checkmark  &89.29& 88.43&90.50& 88.79  \\
\cellcolor{green!8}G   (w/ $\mathcal{L}_{\mathrm{ortho}}$) & \cellcolor{green!8}\Checkmark           & \cellcolor{green!8}\Checkmark             &\cellcolor{green!8} \Checkmark       &  \cellcolor{green!8}\textbf{91.51}& \cellcolor{green!8}\textbf{90.31} &\cellcolor{green!8} \textbf{92.54} &\cellcolor{green!8} \textbf{90.46}  \\  \bottomrule
\end{tabular}
\label{tab_ab}
\end{table}

\textbf{Component analysis.} To systematically quantify the contribution of each architectural component within the ERIS framework, a comprehensive ablation study was conducted, with results presented in Table \ref{tab_ab}. The analysis establishes that a baseline model incorporating the core mechanisms of DSE, LSE, and AR provides a strong foundation for performance. This foundational model (Configuration F), which excludes the orthogonality constraint, achieves a competitive accuracy of 89.29\%. However, the study reveals that the weight-level orthogonality loss, $\mathcal{L}_\mathrm{ortho}$, is the pivotal component for elevating the model to its peak performance. Its criticality is demonstrated by the consistent and significant performance improvements observed across multiple architectural configurations. Most notably, its integration into the full architecture (a transition from Configuration F to G) yields a substantial accuracy increase of over 2.2\%. Consistent performance gains across diverse architectures, notably in a simpler non-autoregressive model (transitioning from Configuration C to D), underscore the fundamental importance of enforcing structural independence between the domain and label parameter subspaces. By explicitly penalizing parameter space overlap, $\mathcal{L}_\mathrm{ortho}$ mitigates feature interference at a foundational level, enabling the model to learn a more disentangled and robust representation. Ultimately, the analysis confirms that the complete model (Configuration G), which incorporates this pivotal orthogonality constraint, achieves superior performance across all evaluated metrics, cementing the role of $\mathcal{L}_\mathrm{ortho}$ as an essential component for maximizing the framework's efficacy.

\section{Conclusion}
In this work, we observe that the unguided nature of feature separation leads to overly entangled domain relationships, which hinder the model's ability to learn invariant features. To address this issue, we propose ERIS, a framework that enhances the disentanglement of domain-specific and label-relevant features through three key mechanisms: an energy-guided calibration mechanism, a weight-level orthogonality strategy, and an auxiliary adversarial generalization branch. We theoretically and empirically demonstrate that the orthogonality strategy not only reduces interference among features but also preserves the discriminative power and information integrity of label-relevant representations. Furthermore, we analyze the impact of the weight-level orthogonality strategy across various baselines. Experimental results demonstrate that ERIS significantly outperforms SOTA methods and highlights its effectiveness in alleviating domain and label entanglement. Future work may further investigate the theoretical boundaries of energy-based modeling in the context of temporal uncertainty.


\bibliographystyle{IEEEtran}
\bibliography{Ref}
\end{document}